\documentclass[10pt,twocolumn,letterpaper]{article}
\usepackage[pagenumbers]{cvpr}
\usepackage{graphicx} \graphicspath{{figures/}}
\usepackage{amsmath,amssymb,mathabx,mathtools,amsthm,nicefrac}
\definecolor{cvprblue}{rgb}{0.21,0.49,0.74}
\usepackage[pagebackref,breaklinks,colorlinks,allcolors=cvprblue]{hyperref}
\usepackage[capitalise]{cleveref}
\usepackage[skip=3pt,font=small]{caption}
\usepackage{acronym}
\usepackage{wrapfig}
\usepackage[dvipsnames,svgnames,x11names,table]{xcolor}
\usepackage{enumitem}
\usepackage{xspace}
\usepackage[misc]{ifsym}
\usepackage{booktabs,tabularx,colortbl,multirow,array,makecell,tabularray}
\usepackage{dsfont}

\makeatletter
\DeclareRobustCommand\onedot{\futurelet\@let@token\@onedot}
\def\@onedot{\ifx\@let@token.\else.\null\fi\xspace}
\def\eg{\emph{e.g}\onedot}

\makeatother

\makeatletter
\renewcommand{\paragraph}{%
  \@startsection{paragraph}{4}%
  {\z@}{0ex \@plus 0ex \@minus 0ex}{-1em}%
  {\hskip\parindent\normalfont\normalsize\bfseries}%
}
\makeatother

\acrodef{sota}[SOTA]{state-of-the-art}
\acrodef{mllm}[MLLM]{Multimodal Large Language Model}
\acrodef{ood}[OOD]{out-of-distribution}
\acrodef{vae}[VAE]{variational autoencoder}
\acrodef{dof}[DoF]{degree of freedom}
\acrodef{fsq}[FSQ]{Finite Scalar Quantization}
\acrodef{fid}[FID]{Fréchet Inception Distance}
\acrodef{mmdist}[MM-dist]{multimodal distance}
\acrodef{rmse}[RMSE]{root mean square error}
\newcommand{\model}{UniAct\xspace}
\newcommand{\dataset}{UA-Net\xspace}

\def\paperID{*****}
\def\confName{CVPR}

\title{UniAct: Unified Motion Generation and Action Streaming for Humanoid Robots\vspace{-12pt}}

\author{
    Nan Jiang\textsuperscript{1,2,3,9,11}\footnotemark[1] \quad
    Zimo He\textsuperscript{4,2,9,11}\footnotemark[1] \quad
    Wanhe Yu\textsuperscript{1,3,5,9,11} \quad
    Lexi Pang\textsuperscript{1,3,5,9,11} \vspace{3pt}\\
    Yunhao Li\textsuperscript{6,3,9,11} \quad
    Hongjie Li\textsuperscript{7,3,9,11} \quad
    Jieming Cui\textsuperscript{1,2,3,9,11} \quad
    Yuhan Li\textsuperscript{8,2} \vspace{3pt}\\
    Yizhou Wang\textsuperscript{4,9,10} \quad
    Yixin Zhu\textsuperscript{3,1,9,11,12}$^{\,\textrm{\Letter}}$ \quad
    Siyuan Huang\textsuperscript{2,9}$^{\,\textrm{\Letter}}$
    \vspace{3pt}\\
    \fontsize{8}{8}\selectfont \textsuperscript{1} Institute for AI, Peking University \quad
    \fontsize{8}{8}\selectfont \textsuperscript{2} Beijing Institute for General Artificial Intelligence (BIGAI) \\
    \fontsize{8}{8}\selectfont \textsuperscript{3} School of Psychological and Cognitive Sciences, Peking University \quad
    \fontsize{8}{8}\selectfont \textsuperscript{4} School of Computer Science, Peking University \\ 
    \fontsize{8}{8}\selectfont \textsuperscript{5} Yuanpei College, Peking University \quad 
    \fontsize{8}{8}\selectfont \textsuperscript{6} School of Foreign Languages, Peking University \quad
    \fontsize{8}{8}\selectfont \textsuperscript{7} School of EECS, Peking University \\
    \fontsize{8}{8}\selectfont \textsuperscript{8} Huazhong University of Science and Technology \quad
    \fontsize{8}{8}\selectfont \textsuperscript{9} State Key Lab of General AI \quad
    \fontsize{8}{8}\selectfont \textsuperscript{10} Nat'l Eng. Research Center of Visual Technology \\
    \fontsize{8}{8}\selectfont \textsuperscript{11} Beijing Key Laboratory of Behavior and Mental Health, Peking University \\
    \fontsize{8}{8}\selectfont \textsuperscript{12} Embodied Intelligence Lab, PKU-Wuhan Institute for Artificial Intelligence \\
    \fontsize{8}{8}\selectfont \footnotemark[1]\;\;Equal contribution \quad 
    $\textrm{\Letter}$\,\,\texttt{yixin.zhu@pku.edu.cn,\,syhuang@bigai.ai}
    \vspace{3pt}\\
    \href{https://jnnan.github.io/uniact/}{https://jnnan.github.io/uniact/}
    \vspace{-18pt}
}

\begin{document}

\twocolumn[{%
    \renewcommand\twocolumn[1][]{#1}%
    \maketitle
    \begin{center}
        \centering
        \captionsetup{type=figure}
        \includegraphics[width=\linewidth]{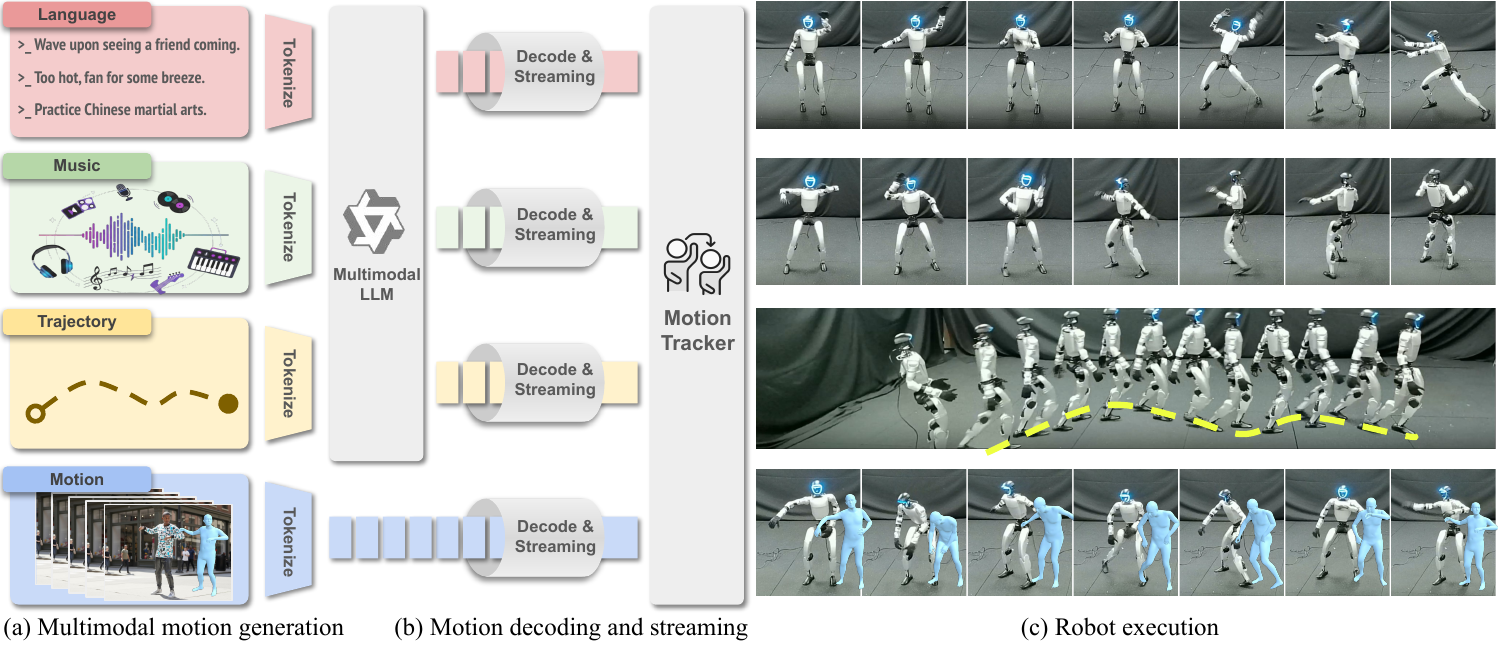} 
        \captionof{figure}{\textbf{\model, a unified framework for multimodal motion generation and action streaming.} \model enables humanoid robots to interpret and execute diverse multimodal instructions---including natural language, musical rhythms, spatial trajectories, and reference motions---with high-fidelity performance. The architecture consists of three core components: (a) a fine-tuned \acs{mllm} that translates heterogeneous inputs into discrete motion tokens via a shared codebook using \acs{fsq}; (b) a causal decoding and streaming pipeline that ensures low-latency delivery of reference motions; and (c) a robust motion tracker that executes the generated motions while maintaining dynamic balance.}
        \label{fig:teaser}
    \end{center}%
}]

\begin{abstract}
A long-standing objective in humanoid robotics is the realization of versatile agents capable of following diverse multimodal instructions with human-level flexibility.
Despite advances in humanoid control, bridging high-level multimodal perception with whole-body execution remains a significant bottleneck. 
Existing methods often struggle to translate heterogeneous instructions---such as language, music, and trajectories---into stable, real-time actions.
Here we show that \model, a two-stage framework integrating a fine-tuned \ac{mllm} with a causal streaming pipeline, enables humanoid robots to execute multimodal instructions with sub-500~ms latency.
By unifying inputs through a shared discrete codebook via \ac{fsq}, \model ensures cross-modal alignment while constraining motions to a physically grounded manifold. This approach yields a 19\% improvement in the success rate of zero-shot tracking of imperfect reference motions.
We validate \model on \dataset, our 20-hour humanoid motion benchmark, demonstrating robust generalization across diverse real-world scenarios.
Our results mark a critical step toward responsive, general-purpose humanoid assistants capable of seamless interaction through unified perception and control.
\end{abstract}

\section{Introduction}

Advances in humanoid control have yielded remarkable progress in low-level tracking~\cite{liao2025beyondmimic,yin2025unitracker}, force control~\cite{zhi2025learning}, and agile locomotion~\cite{long2025learning,sun2025learning}. These developments enable robots to execute complex physical tasks with unprecedented precision. However, a fundamental gap persists between high-level perception and low-level execution, limiting humanoid systems' ability to translate multimodal instructions into coherent, stable actions seamlessly.

Existing attempts to bridge this perception-control gap generally follow two bifurcated paradigms: one-step end-to-end mapping or two-step hierarchical pipelines. One-step methods~\cite{shao2025langwbc,xue2025leverb}, often utilizing \acp{vae} or diffusion models, offer low latency but typically struggle with the long-term temporal dependencies and cross-modal reasoning required for complex instructions. Conversely, two-step methods~\cite{he2024omnih2o,mao2024learning} decouple motion generation from execution; while this improves instruction comprehension, it introduces significant computational overhead and real-time planning challenges. Furthermore, both paradigms remain brittle when encountering \ac{ood} observations or imperfect human demonstrations, often generating physically infeasible motions that result in hardware instability or falls.

The emergence of \acp{mllm}~\cite{luo2024m,zhang2024large,zhou2023ude} and robust motion trackers~\cite{liao2025beyondmimic,he2024omnih2o} provides a promising pathway to resolve this tension. \acp{mllm} excel at reasoning across diverse modalities, while modern trackers can execute reference motions with high fidelity. Leveraging these breakthroughs, we introduce \model, a unified framework that harmonizes high-level multimodal perception with low-level whole-body control via a decoupled two-stage architecture. 

In the first stage, a fine-tuned \ac{mllm} processes diverse inputs---including language, music, and trajectories---to generate discrete motion tokens via a unified codebook. These tokens are then streamed to the second stage, where a causal decoder translates them into real-time commands for a robust motion tracker. This design introduces three key innovations: (i) a unified token representation that facilitates seamless cross-modal translation, (ii) a next-token prediction paradigm that minimizes response latency, and (iii) a discrete action space that restricts generation to a physically grounded manifold, inherently improving tracking robustness against \ac{ood} inputs.

Our evaluation demonstrates \model's versatility across a spectrum of tasks. The system accurately interprets complex sequential commands, synchronizes expressive dance movements with real-time musical beats, and follows precise spatial trajectories while maintaining natural gaits. Notably, our framework achieves a 19\% improvement in zero-shot tracking of low-quality reference motions compared to \ac{sota} baselines. These capabilities were validated through more than 1,000 simulation trials and 100+ hours of real-world operation on physical humanoid platforms.

To advance reproducible research on humanoid control, we contribute \dataset, a comprehensive 20-hour dataset of high-quality robot motions. Meticulously organized in lexicographic order and annotated in multiple modalities, including linguistic descriptions and rhythmic patterns, \dataset serves as a rigorous benchmark for evaluating the multimodal instruction following capabilities.

In summary, our key contributions are:
\begin{itemize}
    \item We present \model, a unified framework that integrates \acp{mllm} with robust whole-body tracking, achieving a response latency of sub-500~ms for multimodal humanoid control.
    \item We introduce a unified motion embedding method that leverages shared discrete tokens to enable seamless cross-modal translation and enhanced tracking stability.
    \item We demonstrate the robust generalization of \model through extensive real-world experiments across diverse linguistic, rhythmic, and trajectory following tasks.
    \item We establish \dataset, a large-scale multimodal benchmark and standardized evaluation protocol to facilitate future research on humanoid embodied intelligence.
\end{itemize}

\section{Related work}

\paragraph{Whole-body humanoid control}
Humanoid control has its foundations in physics-based animation, where reinforcement learning is extensively used to track reference motions with high fidelity~\cite{peng2018deepmimic,luo2023perpetual,serifi2024vmp,peng2021amp}. To facilitate high-level task specification, research has introduced adversarial embeddings~\cite{peng2022ase,cui2024anyskill} and task-specific latent policies~\cite{tessler2024maskedmimic,pan2025tokenhsi} to enable behavioral synthesis through compositional inputs~\cite{xu2023composite,wang2025skillmimic,tirinzoni2025zero,chen2024taming}. Recent approaches further integrate diffusion-based motion planning with robust tracking policies for multi-task character control~\cite{tevet2024closd}. Despite their effectiveness in simulation, these animation-centric methods fundamentally rely on privileged observations that are typically unavailable in real-world settings.

In the physical domain, humanoid control has progressed through robust low-level mechanisms~\cite{liao2025beyondmimic,yin2025unitracker,he2024omnih2o,fu2024humanplus,zhang2024wococo,he2025hover,ji2024exbody2,ze2025twist,li2025amo,he2025asap} capable of handling diverse modalities, including directional locomotion~\cite{radosavovic2024real,bloesch2022towards,duan2021learning,sun2025learning,radosavovic2024learning}, end-effector pose specification~\cite{li2025clone,he2024omnih2o}, and natural language commands~\cite{xue2025leverb,shao2025langwbc,li2025language}. However, existing frameworks generally treat these modalities in isolation through distinct injection mechanisms~\cite{xue2025leverb,ding2025humanoid}. Language-conditioned approaches, in particular, face a stark trade-off: hierarchical pipelines~\cite{he2024omnih2o,he2024learning} often sacrifice real-time responsiveness for comprehension, while end-to-end architectures that incorporate language latents directly~\cite{shao2025langwbc,li2025language} often struggle with complex semantic reasoning. Consequently, the field lacks a unified framework capable of processing heterogeneous multimodal inputs while maintaining deterministic, real-time control.

\paragraph{Multimodal human motion generation}
Motion generation has evolved from specialized text-driven approaches~\cite{plappert2018learning,text2action,dvgans,seq2seq,fan2025go,jl2p,humanml3d,motionclip,temos,avatarclip,motiondiffuse,mdm,mld,t2mgpt,tmr,humantomato} to versatile systems incorporating diverse control modalities, such as spatial trajectories~\cite{gmd,lu2022action,omnicontrol,pinyoanuntapong2025maskcontrol} and rhythmic musical signals~\cite{le2023music,tseng2023edge,sun2022you,li2023finedance,alexanderson2023listen,valle2021transflower}. Recent unified frameworks~\cite{luo2024m,zhang2024large,zhou2023ude,bian2025motioncraft,li2024mulsmo,li2025omnimotion} seek to harmonize these modalities within a single architecture: M$^3$GPT~\cite{luo2024m} employs discrete vector quantization for cross-modal synthesis, UDE~\cite{zhou2023ude} utilizes modality-agnostic transformers, and MotionCraft~\cite{bian2025motioncraft} adopts a coarse-to-fine training strategy for varying control granularities. While these methods achieve impressive offline generation quality, they predominantly rely on iterative diffusion processes or multi-stage pipelines that introduce substantial computational latency. Consequently, they remain ill-suited for humanoid control applications that demand instantaneous, real-time response.

\paragraph{Humanoid motion dataset}
Progress in humanoids is significantly hampered by the scarcity of high-quality datasets compared to fixed-base manipulation platforms~\cite{fang2025ai,brohan2022rt,walke2023bridgedata,driess2023palm}. To bridge this gap, motion retargeting techniques~\cite{luo2023perpetual,zakka10mink,ze2025gmr} have been developed to adapt human MoCap data from general-purpose repositories~\cite{harvey2020robust,mahmood2019amass,li2023object} to humanoid morphologies. Specialized datasets like OmniH2O~\cite{he2024omnih2o}, OmniRetarget~\cite{yang2025omniretarget}, and TWIST~\cite{ze2025twist} have further advanced the field by providing loco-manipulation data tailored to specific robot embodiments. However, existing datasets focus on physical trajectories while neglecting the underlying semantic meanings and the need for multimodal alignment. This lack of comprehensive, semantically rich benchmarks constrains the development of humanoids capable of following multifaceted, complex instructions.

\section{The \model model}

\model addresses the challenge of translating diverse multimodal instructions into robust humanoid motion through a unified three-stage pipeline, as illustrated in \cref{fig:teaser}. The framework consists of: (i) \acs{fsq}-based instruction tokenization that unifies heterogeneous inputs~\cite{mentzer2023finite}, (ii) \acs{mllm}-driven motion generation that produces temporally coherent motion sequences, and (iii) real-time whole-body tracking that executes generated motions on physical hardware.

\begin{figure}[t!]
    \centering
    \includegraphics[width=\linewidth]{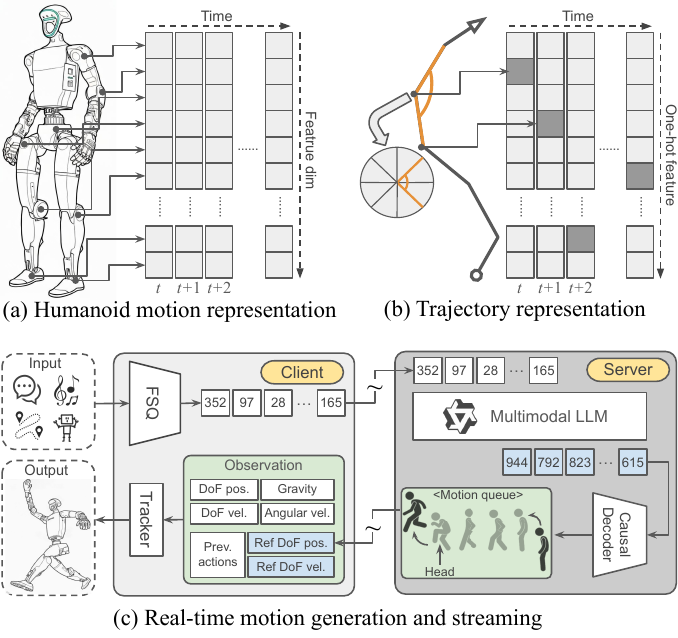}
    \caption{\textbf{Overview of \model and multimodal representations.} 
    (a) Humanoid motion is represented as temporal sequences of \acs{dof} positions, tokenized via \acs{fsq}. 
    (b) Trajectory features are one-hot encoded based on the turning angle degree of segmented paths. 
    (c) System architecture: the server-side \acs{mllm} processes multimodal inputs (text, music, trajectory) and \acs{fsq}-tokenized motions to autoregressively generate motion tokens; a causal decoder transforms tokens to continuous \acs{dof} positions, which are streamed to the client and executed by the tracking controller for real-time motion synthesis.}
    \label{fig:method}
\end{figure}

As detailed in \cref{fig:method}, the system first encodes text, music, trajectories, and reference motions into a shared discrete token representation. A fine-tuned \ac{mllm} then processes these tokens to autoregressively generate motion sequences. Finally, a causal decoder streams the generated tokens to a robust motion tracker, enabling immediate execution on the humanoid platform with sub-500~ms latency.

\subsection{Problem formulation}

We formulate the multimodal humanoid control problem as follows. Given a multimodal instruction $\mathcal{C} = \{\mathcal{C}_\text{text}, \mathcal{C}_\text{music}, \mathcal{C}_\text{traj}, \mathcal{C}_\text{motion}\}$, where each component represents an optional control signal, \model generates a temporal sequence of target \ac{dof} positions $\mathbf{p}_t \in \mathbb{R}^{D}$ at each control timestep $t$, with $D$ denoting the robot's total \ac{dof}. The key challenge lies in bridging the semantic gap between high-level multimodal instructions and low-level joint-space commands while maintaining real-time performance and physical feasibility.

\subsection{Multimodal instruction tokenization}

To enable unified processing across heterogeneous modalities, we map continuous input signals $\mathbf{X}$ into a discrete latent space compatible with autoregressive language modeling. Our tokenization strategy adapts to the distinct physical and semantic characteristics of each modality while maintaining a shared discrete vocabulary.

\paragraph{Text}
Text instructions are tokenized using the native vocabulary of Qwen2.5~\cite{bai2025qwen25vltechnicalreport}, directly leveraging the model's pre-trained word embeddings and linguistic understanding capabilities. Unlike other modalities described below, text requires no additional encoding.

\paragraph{Music}
Music signals are processed at 30~Hz following the AIST++~\cite{li2021ai} protocol to extract temporal features. At each timestep, we compute a 35-dimensional feature vector comprising: the envelope (capturing amplitude variations), 20 MFCC coefficients (representing timbral characteristics), 12 chroma features (encoding harmonic content), and binary indicators for beat peaks and onsets. This representation captures both rhythmic structure and timbral qualities essential for dance generation.

\paragraph{Trajectory}
Trajectories are represented as the angles of line segments (\cref{fig:method}(b)). We focus on heading direction variations to capture locomotion patterns while maintaining computational efficiency. The continuous trajectory is segmented at 5~FPS; at each frame $i$, we compute the root displacement in the previous frame's coordinate system:
\begin{equation}
    \mathbf{r}_i = \mathbf{R}_{i-1}^\mathrm{T} (\mathbf{p}_i - \mathbf{p}_{i-1}) \in \mathbb{R}^3,
\end{equation}
where $\mathbf{p}_i$ denotes the root position and $\mathbf{R}_{i-1}$ is the rotation matrix at frame $i-1$. This formulation ensures orientation-invariant encoding. The angular difference in heading direction is then discretized into 6-degree bins, yielding a compact codebook of size $360/6=60$.

\paragraph{Motion}
Reference motions are represented directly as humanoid \ac{dof} positions, eliminating the need for retargeting and enabling end-to-end control. For the Unitree-G1 robot with 29 \acp{dof}, each frame is encoded as $\mathbf{q}_i \in \mathbb{R}^{29}$ containing all joint angle values.

To unify music, trajectory, and motion representations into discrete tokens compatible with autoregressive language modeling, we employ \ac{fsq}. For music and motion modalities, we train separate encoder-decoder pairs that compress continuous features into latent representations, which are then element-wise quantized:
\begin{equation}
    \text{FSQ}(\mathbf{z}) = \text{round}(\text{tanh}(\mathbf{z}) \cdot \mathbf{L}),
\end{equation}
where $\mathbf{z}$ denotes the encoder output and $\mathbf{L}$ specifies the quantization levels per dimension. Both encoder and decoder are implemented as 1D convolutional networks with residual connections, trained to minimize reconstruction loss $\mathcal{L}_\text{FSQ} = ||\mathbf{X} - \hat{\mathbf{X}}||_2^2$ (details in \cref{sec:supp:implementation_tokenization}). This architecture preserves temporal coherence while maintaining computational efficiency. The quantized latents are subsequently mapped to discrete token indices, creating modality-specific vocabularies $\mathcal{V}_\text{music}$, $\mathcal{V}_\text{traj}$, and $\mathcal{V}_\text{motion}$.

\subsection{Motion generation}

We fine-tune Qwen2.5-3B~\cite{bai2025qwen25vltechnicalreport} to autoregressively predict motion tokens conditioned on multimodal inputs. Our approach unifies all modalities through early fusion, constructing a shared vocabulary by concatenating tokens:
\begin{equation}
    \mathcal{V}_\text{unified} = \mathcal{V}_\text{text} \cup \mathcal{V}_\text{music} \cup \mathcal{V}_\text{traj} \cup \mathcal{V}_\text{motion}.
\end{equation}
To maintain a manageable vocabulary size, we substitute the least frequent text tokens with tokens from other modalities. This substitution preserves the model's capacity for common linguistic instructions while accommodating the additional modality-specific vocabularies.

\paragraph{Sequence structure and training}
During training, we wrap each modality's token sequence with special delimiter tokens that clearly delineate modality boundaries. The complete input sequence takes the form of a concatenation of all available modalities---each enclosed by its respective delimiters---followed by the target motion sequence. This explicit boundary marking allows the model to distinguish between different conditioning modalities and the motion tokens to be predicted.

To support generation with arbitrary-length conditions, we apply a sliding window approach to trajectory and music modalities during training. Specifically, we randomly sample fixed-length segments from the full token sequences of these modalities. This approach enables the model to learn from diverse temporal contexts while maintaining computational efficiency through bounded sequence lengths.

The model is optimized via standard autoregressive language modeling:
\begin{equation}
    \mathcal{L}_\text{gen} = -\sum_{t=1}^{T} \log P(\mathbf{s}_t^\text{motion} | \mathbf{s}_{<t}, \mathcal{C}),
\end{equation}
where $\mathbf{s}_{<t}$ represents all preceding tokens (including both conditioning modalities and previously generated motion tokens), and $\mathcal{C}$ denotes the multimodal instruction context.

\paragraph{Inference}
During inference, the model autoregressively generates motion tokens conditioned on the input modalities. The generated discrete tokens are then decoded back to continuous \ac{dof} positions through the trained \ac{fsq} decoder, producing target reference motions for the humanoid tracking controller.

\begin{figure*}[t!]
    \centering
    \includegraphics[width=\linewidth]{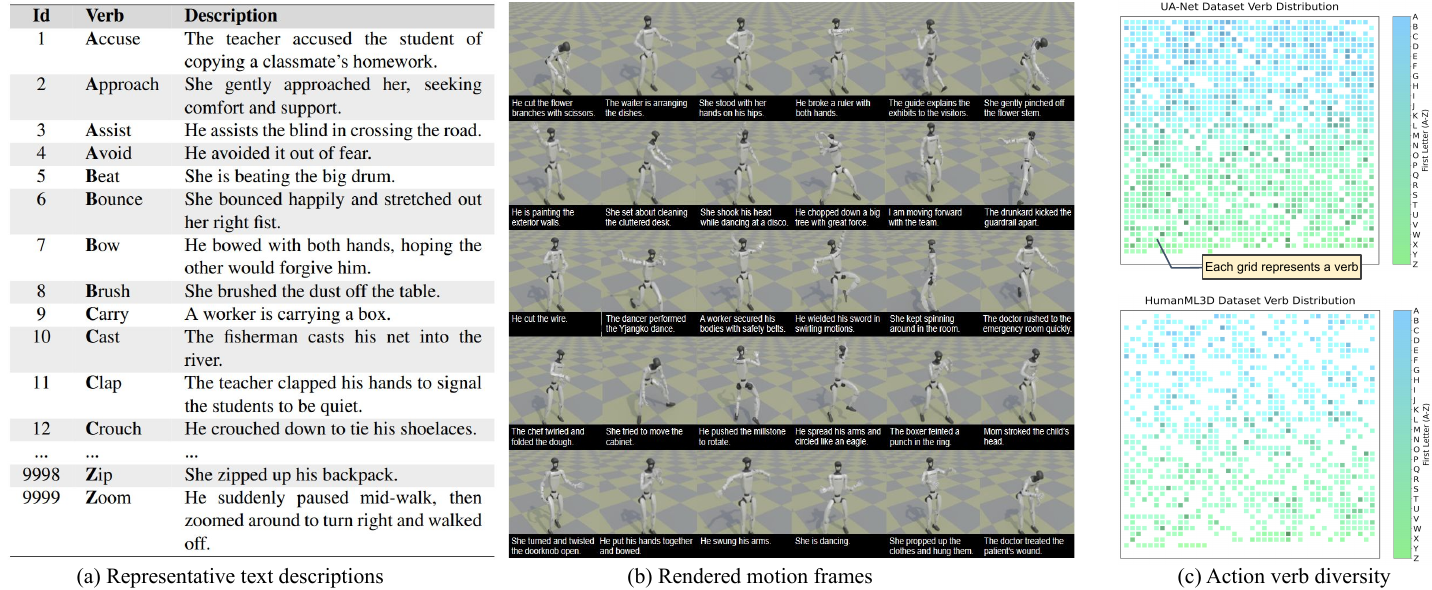}
    \caption{\textbf{\dataset dataset analysis.} 
    (a) Representative text descriptions of human motions from \dataset. 
    (b) Rendered motion frames corresponding to selected descriptions. 
    (c) Action verb diversity comparison: we visualize the presence of 1684 common verbs on a square grid organized alphabetically (A--Z), where each cell represents a verb. \dataset demonstrates significantly broader vocabulary coverage compared to HumanML3D~\cite{humanml3d}. (Vector graphics; zoom for details.)}
    \label{fig:dataset}
\end{figure*}

\subsection{Causal motion decoding}

To enable real-time motion execution, we implement a causal decoding mechanism that transforms generated motion tokens into humanoid \ac{dof} positions in a streaming fashion. The causal constraint ensures that the decoded motion at timestep $t$ depends only on tokens up to time $t$, enabling online inference without future information:
\begin{equation}
    \mathbf{h}_t = \sum_{k=0}^{K-1} \mathbf{W}_k \cdot \mathbf{h}_{t-k} + \mathbf{b},
\end{equation}
where $K$ denotes the kernel size, and $\mathbf{W}_k$ and $\mathbf{b}$ are learned convolution parameters. To balance computational efficiency with low latency, we adopt a chunked decoding strategy where decoding is triggered only when accumulated tokens reach a predefined chunk size, reducing overhead while maintaining responsive motion generation.

\subsection{Real-time motion streaming}

We implement a server-client streaming architecture that decouples computationally intensive motion generation (server-side) from real-time robot control (client-side). When instructions are provided through the client interface, the commands are forwarded to the GPU-equipped server, which tokenizes the multimodal instruction and feeds it to the Qwen2.5 model for autoregressive motion generation. Generated tokens are continuously decoded and streamed to the client via low-latency WebSocket communication. To ensure smooth transitions between consecutive instructions, we maintain 10 motion tokens as history context when processing new commands.

The server generates motions faster than real-time, creating a temporal mismatch between the server's variable generation rate and the client's fixed 50~Hz control frequency. We address this through a client-side motion cache that accumulates incoming frames from the server. The client queries this buffer at precise 20~ms intervals, ensuring consistent reference motion delivery to the tracking controller regardless of server-side generation speed variations. This decoupling guarantees smooth playback even when the generation experiences momentary slowdowns or speedups.

\subsection{Humanoid motion tracking}

The decoded motion sequences are streamed to a modified BeyondMimic~\cite{liao2025beyondmimic} tracker that serves as the low-level controller. We adapt the original tracker by removing the global orientation component from the reference motion observation, allowing the policy to focus exclusively on relative joint configurations rather than absolute orientation:
\begin{equation}
    \mathbf{o}_t = [\mathbf{q}_t^\text{ref}, \dot{\mathbf{q}}_t^\text{ref}, \mathbf{q}_t^\text{curr}, \dot{\mathbf{q}}_t^\text{curr}, \mathbf{g}_t, \mathbf{\omega}_t, \mathbf{a}_{t-1}],
\end{equation}
where $\mathbf{q}_t^\text{ref}$ represents the decoded reference joint positions, $\mathbf{q}_t^\text{curr}$ denotes the current robot joint state, $\mathbf{g}_t$ is the gravity vector in the robot's body frame, $\mathbf{\omega}_t$ is the angular velocity, and $\mathbf{a}_{t-1}$ is the previous action. 

The tracker outputs target joint positions $\mathbf{p}_t$ that drive the robot to follow the generated motion while maintaining dynamic balance and physical feasibility (see \cref{sec:supp:motion-tracking} for training details). By decoupling high-level motion generation from low-level control, this architecture enables robust execution even under external disturbances, as the tracker can reactively adjust joint commands while preserving the overall motion intent.

\section{The \dataset dataset}

We present \dataset, a comprehensive multimodal humanoid motion dataset comprising text-to-motion, trajectory-to-motion, and music-to-motion modalities. \dataset combines our collected MoCap data for text and trajectory conditioning with the existing FineDance dataset~\cite{li2023finedance} for music. All human motions are retargeted to humanoid-compatible formats using GMR~\cite{ze2025gmr}, with manual refinement to preserve motion fidelity while respecting humanoid physical constraints (see \cref{sec:supp:dataset} for details).

\begin{table*}[t!]
    \centering
    \small
    \setlength{\tabcolsep}{4pt}
    \caption{\textbf{Multimodal humanoid control evaluation.} We compare \model against baselines across text, trajectory, and music modalities. Metrics include \acs{fid} (motion quality), Diversity (motion variation), \acs{mmdist} (text-motion alignment), R-precision (retrieval accuracy at top-1/2/3), Root error (trajectory tracking accuracy in meters), Genre (genre fidelity for dance), and Success rate (task completion without falls). Arrows indicate better direction. We ablate \acs{fsq} codebook size (0.25× codes) and downsampling rate (2× downsampled). \model achieves superior performance across most metrics while maintaining lower diversity than diffusion-based two-step methods.}
    \label{tab:multimodal_control}
    \resizebox{\linewidth}{!}{%
        \begin{tabular}{llccccccccc}
            \toprule
            \multirow{2}{*}{Modality} & \multirow{2}{*}{Method} & \multirow{2}{*}{\acs{fid} $\downarrow$} & \multirow{2}{*}{Diversity $\uparrow$} & \multirow{2}{*}{\acs{mmdist} $\downarrow$} & \multicolumn{3}{c}{R-precision $\uparrow$} & \multirow{2}{*}{Root error $\downarrow$} & \multirow{2}{*}{Genre $\uparrow$} & \multirow{2}{*}{Success rate $\uparrow$} \\
            \cmidrule(lr){6-8}
            & & & & & R@1 & R@2 & R@3 & & & \\
            \midrule
            \multirow{7}{*}{{Text}} 
            & LangWBC~\cite{shao2025langwbc} & 5.17 & 5.03 & 4.19 & 10.21 & 14.47 & 19.73 & - & - & 58.2 \\
            & UH-1~\cite{mao2024learning} & 6.39 & 5.41 & 4.45 & 12.73 & 14.30 & 17.23 & - & - & 51.3 \\
            & OmniH2O~\cite{he2024omnih2o}+MDM~\cite{mdm} & 3.11 & \textbf{5.88} & 3.29 & 18.01 & 25.42 & 27.94 & - & - & 63.0 \\
            & BeyondMimic~\cite{liao2025beyondmimic}+MDM~\cite{mdm} & 2.93 & 5.72 & 2.93 & 24.53 & 28.78 & 34.68 & - & - & 65.3 \\
            & Ours  & \textbf{1.69} & 5.21 & \textbf{2.45} & \textbf{41.59} & \textbf{54.28} & \textbf{60.63} & - & - & \textbf{83.1} \\
            & Ours (0.25× codes) & 1.84 & 5.04 & 2.67 & 33.19 & 36.42 & 38.75 & - & - & 79.6 \\
            & Ours (2× downsamp.) & 1.73 & 4.92 & 2.51 & 38.03 & 50.90 & 58.13 & - & - & 80.9 \\
            \midrule
            \multirow{5}{*}{{Trajectory}} 
            & OmniH2O~\cite{he2024omnih2o}+MDM~\cite{mdm} & 2.62 & 3.42 & 2.75 & 11.41 & 19.06 & 24.74 & 1.392 & - & 23.6 \\
            & BeyondMimic~\cite{liao2025beyondmimic}+MDM~\cite{mdm} & 2.58 & \textbf{3.53} & 2.65 & 12.33 & 17.23 & 27.16 & 1.284 & - & 35.2 \\
            & Ours  & \textbf{0.77} & 3.01 & \textbf{1.77} & \textbf{56.15} & \textbf{65.26} & \textbf{71.43} & \textbf{0.151} & - & \textbf{97.3} \\
            & Ours (0.25× codes) & 0.80 & 3.02 & 1.78 & 55.60 & 60.32 & 65.89 & 0.195 & - & 94.3 \\
            & Ours (2× downsamp.) & 0.79 & 3.13 & 1.82 & 54.18 & 63.15 & 69.90 & 0.163 & - & 95.8 \\
            \midrule
            \multirow{5}{*}{{Music}} 
            & OmniH2O~\cite{he2024omnih2o}+MDM~\cite{mdm} & 2.35 & 4.50 & 3.59 & 15.64 & 21.27 & 23.43 & - & 0.84 & 45.8 \\
            & BeyondMimic~\cite{liao2025beyondmimic}+MDM~\cite{mdm} & 1.97 & \textbf{4.66} & 3.37 & 17.91 & 21.20 & 26.59 & - & 0.77 & 57.1 \\
            & Ours  & \textbf{1.53} & 4.34 & \textbf{2.61} & \textbf{53.13} & \textbf{65.05} & \textbf{69.82} & - & \textbf{0.97} & \textbf{87.4} \\
            & Ours (0.25× codes) & 1.74 & 4.42 & 2.79 & 48.10 & 61.92 & 66.86 & - & 0.88 & 75.6 \\
            & Ours (2× downsampled) & 1.58 & 4.31 & 2.84 & 45.63 & 58.99 & 66.91 & - & 0.80 & 84.4 \\
            \bottomrule
        \end{tabular}%
    }%
\end{table*}

\subsection{Dataset modalities}

\textbf{Text-to-motion} contains 20 hours of motion paired with natural language descriptions, ranging from simple atomic actions to complex compositional movements (\cref{fig:dataset}). Each sequence is organized in lexicographic order based on its primary description, creating a structured taxonomy from basic locomotion (\eg `backward stepping,' `forward walking') to complex interactions (\eg `cooking,' `door opening'). \cref{fig:dataset}(c) shows that \dataset provides broader coverage of common verbs compared to HumanML3D~\cite{humanml3d}, critical for robot motion training and evaluation.

\textbf{Trajectory-to-motion} comprises 20 minutes of walking conditioned on spatial trajectories, including point-to-point navigation, curved paths, and obstacle avoidance. Speed variations span from 0.3~m/s to 1.5~m/s, with annotations specifying velocity, curvature, and heading angle at 120~Hz.

\textbf{Music-to-motion} incorporates 376 retargeted dance sequences from FineDance~\cite{li2023finedance}, spanning multiple genres and tempos. Retargeting preserves rhythmic synchronization while adapting movements to humanoid kinematic and dynamic constraints.

\subsection{MoCap configuration}

We employ an OptiTrack MoCap with 48 high-speed cameras operating at 120~Hz in a 10×8×4~meter capture volume. Data collection uses a 43-marker configuration with additional hand and foot markers for fine-grained MoCap. We recruited 10 professional performers, including dancers, athletes, and motion specialists, to ensure diverse styles.

\begin{figure*}[t!]
    \centering
    \includegraphics[width=\linewidth]{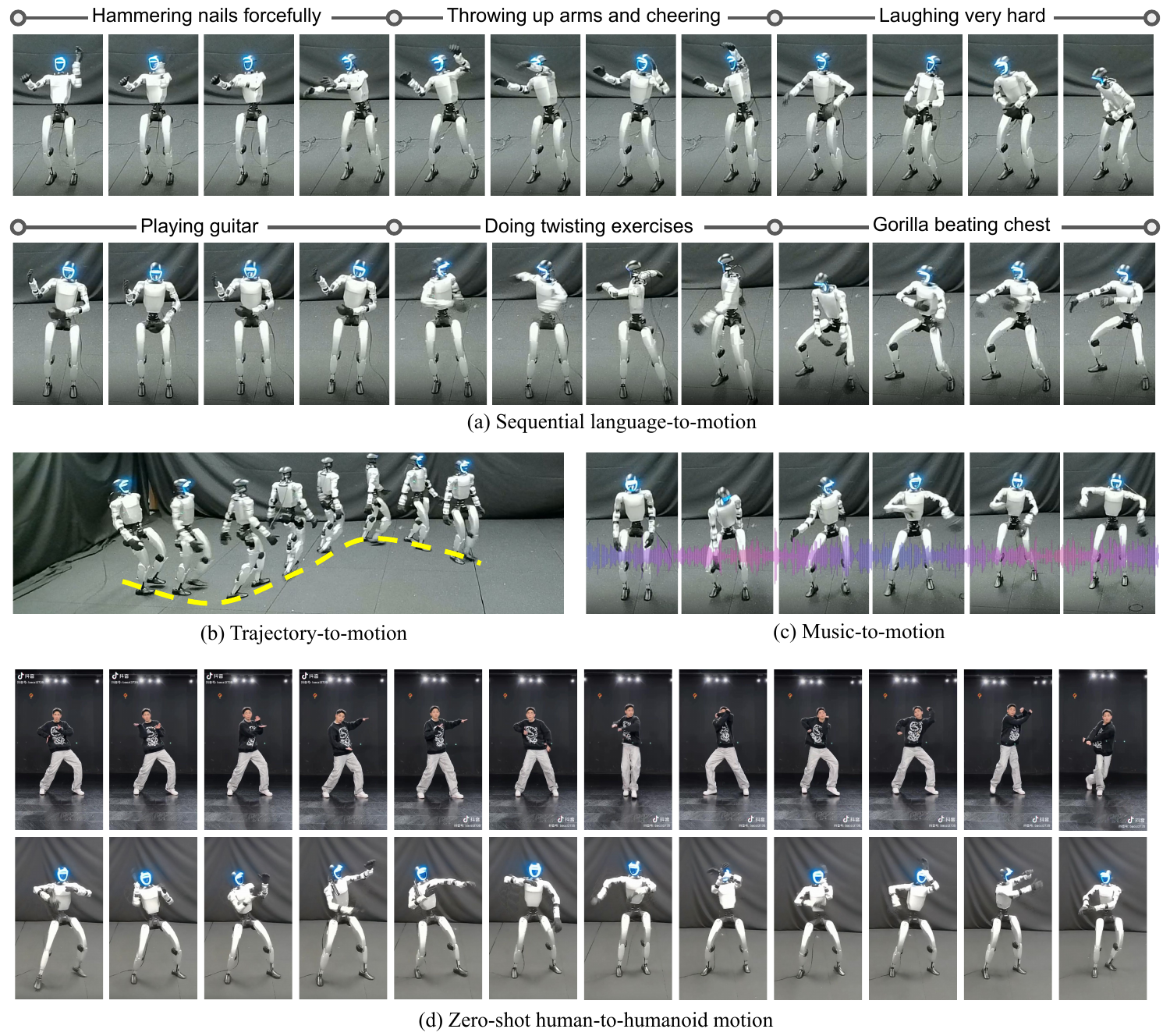}
    \caption{\textbf{Qualitative results of \model across diverse instruction modalities.} (a) Sequential text-to-motion: the humanoid executes a sequence of complex actions following instructions. (b) Trajectory-to-motion: the robot follows a curved path (yellow dashed line) with natural walking motions. (c) Music-to-motion: the humanoid generates dance movements synchronized to the music's rhythm. (d) Zero-shot human-to-humanoid motion transfer: retarget motions from internet videos to humanoid execution without additional training.}
    \label{fig:qualitative}
\end{figure*}

The GMR~\cite{ze2025gmr} pipeline maps captured motions to a 29-\acs{dof} humanoid model compatible with Unitree G1, incorporating collision avoidance, joint limits, and balance constraints. All sequences undergo manual verification to validate foot contact preservation, center of mass trajectories, and dynamic feasibility.

\section{Experiments and results}\label{sec:exp}

We conduct comprehensive experiments to validate \model across three key dimensions: (i) multimodal humanoid control that surpasses strong baselines, (ii) real-time system performance through detailed timing analysis, (iii) motion tokenizer robustness under disturbances and \ac{ood} motions.

\subsection{Multimodal humanoid control}
We evaluate \model's ability to control humanoid robots through diverse input modalities: text, trajectory, and music. From \dataset, we exclude motion sequences involving sitting on chairs, lying down, or stair climbing to ensure physical feasibility in our evaluation environment. The dataset is partitioned with 80\% for training and 20\% for evaluation. Each experiment involves providing a single instruction and evaluating the resulting motion quality.

\paragraph{Experimental protocol} 
For text-to-motion, we provide single sentences from the test set as input, allowing the humanoid to execute actions until reaching the end-of-motion token. For trajectory-to-motion, we select 10-second walking trajectory segments from the test set and autoregressively drive the humanoid to follow the specified path. For music-to-motion, we randomly sample 10-second music clips from FineDance.

\paragraph{Strong baselines and ablations}
We compare \model against several \ac{sota} baselines. LangWBC~\cite{shao2025langwbc} and UH-1~\cite{mao2024learning} directly map text instructions to actions. OmniH2O~\cite{he2024omnih2o} and BeyondMimic~\cite{liao2025beyondmimic} combined with MDM-generated~\cite{mdm} motions serve as two-step methods. Baseline adaptation details are in \cref{sec:supp:baseline_adaptation}. Additionally, we ablate: (i) \acs{fsq} codebook size, and (ii) \acs{fsq} downsampling rates.

\begin{table*}[t!]
    \centering
    \small
    \setlength{\tabcolsep}{3pt}
    \caption{\textbf{Motion tracking evaluation with discrete motion representation.} We evaluate BeyondMimic~\cite{liao2025beyondmimic} tracking performance with and without our \acs{fsq}-based tokenizer across three settings: low-quality reference motions, unseen \acs{ood} motions, and training set reconstruction. Metrics include MPJPE (cm), Error vel. (cm/s), and Success rate. Arrows indicate better direction. Our tokenizer improves robustness on noisy and \acs{ood} motions while maintaining reconstruction quality.}
    \label{tab:motion_tracking}
    \resizebox{\linewidth}{!}{%
        \begin{tabular}{lccccccccc}
            \toprule
            \multirow{2}{*}{Method} & \multicolumn{3}{c}{Low-quality reference motions} & \multicolumn{3}{c}{Unseen motion generation} & \multicolumn{3}{c}{Quantative error analysis} \\
            \cmidrule(lr){2-4} \cmidrule(lr){5-7} \cmidrule(lr){8-10}
             & MPJPE $\downarrow$ & Error vel. $\downarrow$ & Suc. rate $\uparrow$ & MPJPE $\downarrow$ & Error vel. $\downarrow$ & Suc. rate $\uparrow$ & MPJPE $\downarrow$ & Error vel. $\downarrow$ & Suc. rate $\uparrow$ \\
            \midrule
            BeyondMimic~\cite{liao2025beyondmimic}+\acs{fsq} (Ours) & \textbf{36.10} & \textbf{52.62} & \textbf{95.4} & \textbf{45.61} & \textbf{31.60} & \textbf{95.2} & 14.48 & 22.06 & \textbf{100.0} \\
            BeyondMimic~\cite{liao2025beyondmimic}+\acs{fsq} (0.25× codes)  & 50.90 & 61.99 & 90.5 & 51.33 & 37.49 & 91.0 & 23.32 & 25.02 & \textbf{100.0} \\
            BeyondMimic~\cite{liao2025beyondmimic}+\acs{fsq} (2× downsampled)  & 46.19 & 60.29 & 88.4 & 57.78 & 37.44 & 87.2 & 20.07 & 23.10 & \textbf{100.0} \\
            BeyondMimic~\cite{liao2025beyondmimic} & 68.34 & 66.86 & 76.2 & 48.93 & 37.35 & 90.5 & \textbf{12.59} & \textbf{19.99} & \textbf{100.0} \\
            \bottomrule
        \end{tabular}%
    }%
\end{table*}

\paragraph{Evaluation metrics}
Following \citet{humanml3d}, we measure robot motion execution quality in simulation through \ac{fid}, diversity, \ac{mmdist}, and R-precision. Success rate indicates the percentage of trials where the robot completes tasks without falling or significant instruction deviation. For trajectory-following, we quantify tracking accuracy using \ac{rmse} between generated and target trajectories. For dance generation, we introduce a genre fidelity metric measuring the distance between generated dance motions and each dataset genre. All experiments are conducted in MuJoCo with contact dynamics and actuator models. Metric formulations are in \cref{sec:supp:metric}.

\paragraph{Results}
As \cref{tab:multimodal_control} shows, \model achieves the highest performance across most metrics. For text-to-motion, we demonstrate strong text-motion alignment and high success rates, highlighting \model's ability to generate accurate and robust motions from instructions. For trajectory-following, \model achieves the lowest mean root error, demonstrating superior locomotion capabilities. For music-to-dance, we produce strong genre correspondence, effectively matching dance movements to musical styles. While our diversity scores are lower than two-step methods, this is partly attributed to the inherently high diversity of diffusion-based motion generation, which may produce more varied but potentially less controlled outputs. \cref{fig:qualitative} shows qualitative results of \model across diverse instruction modalities.

\begin{table}[t!]
    \centering
    \small
    \setlength{\tabcolsep}{3pt}
    \caption{\textbf{Computational timing breakdown (ms).} We measure latency components across different hardware configurations. Model latency is the sum of motion generation, token decode, and motion track. Total delay additionally includes data transmission when transitioning to new commands. All methods achieve real-time performance at 20~Hz control frequency.}
    \label{tab:timing_breakdown}
    \resizebox{\linewidth}{!}{%
        \begin{tabular}{lcccccc}
            \toprule
            \multirow{2}{*}{Hardware} & Motion & Token & Motion & Data & Model & Total \\
            & generation & decode & track & transmission & latency & delay \\
            \midrule
            H100     & 14.1 & 0.98 & 3.49 & 33.5 & 52.1 & 264 \\
            RTX 5090 & 18.7 & 1.05 & 6.59 & 33.5 & 59.8 & 340 \\
            RTX 4090 & 26.2 & 1.20 & 7.06 & 33.5 & 68.0 & 461 \\
            \bottomrule
        \end{tabular}%
    }%
\end{table}

\subsection{Computational time analysis}

Real-time performance is critical for \model. We quantify system latency through two metrics: total delay (latency when transitioning to new commands) and model latency (latency during ongoing instruction streaming).

\cref{tab:timing_breakdown} breaks down timing into four components: (i) motion generation, time from instruction receipt to first motion token; (ii) token decoding, conversion from discrete tokens to continuous \acs{dof} positions; (iii) motion tracking, low-level controller computation time; and (iv) data transmission, network latency between server and client.

\subsection{Motion tracking evaluation}

We validate that discrete motion representation enhances tracking robustness without compromising fidelity. Using BeyondMimic~\cite{liao2025beyondmimic}, we track motions after encoding, quantizing, and decoding. Performance is measured via MPJPE, joint velocity error, and success rate across three settings:

\paragraph{Low-quality reference motions}
We add Gaussian noise and temporal jitter to \dataset motions, simulating imperfect inputs. Comparing tracking performance after discrete processing against original unperturbed references tests whether quantization refines noisy inputs (\cref{sec:supp:robustness_analysis}).

\paragraph{Unseen motion generalization}
We collect 50 online videos beyond \dataset's training distribution, extract SMPL-X parameters~\cite{pavlakos2019expressive} via GVHMR~\cite{shen2024world}, and retarget to G1 \acs{dof} via GMR~\cite{ze2025gmr}. Comparing performance with/without discrete representation demonstrates how tokenization constrains motions to physically feasible spaces.

\paragraph{Quantization error analysis}
We evaluate reconstruction error on training motions across different configurations, varying \acs{fsq} codebook size and downsampling rate.

\paragraph{Results}
\cref{tab:motion_tracking} shows our tracker maintains higher success rates on reference motions with jitter or \acs{ood} characteristics without quality degradation, demonstrating \model maps motions into the seen domain for robust control.

\subsection{Compositional cross-modal control}

\model enables cross-modal control by independently generating upper-body actions and lower-body trajectories, then fusing them into coherent whole-body movements. Fine-tuning the tracking policy with composed pairs ensures stable execution. This allows expressive actions (\eg, waving, drumming) while following paths. Zero-shot generalization to unseen action-trajectory combinations enhances behavioral diversity and deployment scalability (\cref{sec:supp:cross-modal-control}).

\section{Conclusion}

We introduce \model, a unified framework for real-time multimodal humanoid control with sub-500~ms response latency. By integrating multimodal large language models with efficient motion tokenization, \model translates diverse inputs---text, vision, and audio---into robust robot motions through a shared discrete token representation. Extensive real-world experiments demonstrate superior performance across multimodal tasks while maintaining robustness to noisy and \acs{ood} inputs. We also establish a comprehensive benchmark and dataset with standardized evaluation protocols for the research community.

\paragraph{Limitation}
The system struggles with highly dynamic movements (\eg, rapid jumping) due to motion tracker constraints, and currently lacks object manipulation capabilities. Future work could develop more robust controllers for high-speed motions and incorporate object-centric representations to enable contact-rich manipulation tasks.

\paragraph{Acknowledgement}
This work is supported in part by the National Science and Technology Innovation 2030 Major Program (2025ZD0219400), the National Natural Science Foundation of China (62376009 to Y.~Zhu and 6247070125 to Y.~Wang), the PKU-BingJi Joint Laboratory for Artificial Intelligence, and the National Comprehensive Experimental Base for Governance of Intelligent Society, Wuhan East Lake High-Tech Development Zone. This project would not have been possible without the year-long collaboration with Virtual Point, whose partnership has been instrumental to this work. We are deeply grateful to Peiyuan Zhi and Le Ma from BIGAI, and Wenhua Xia and Rui Chen from the PKU-Wuhan Institute of Artificial Intelligence for their invaluable hardware support and technical assistance throughout the development process. We also extend our sincere thanks to Yutang Lin and Yixuan Li for their insightful discussions and feedback that helped shape the ideas presented in this work. Finally, we thank all the team members who contributed to the countless hours of physical robot experiments and debugging that made this system a reality.

{
    \small
    \bibliographystyle{ieeenat_fullname}
    \bibliography{reference_header,reference}
}

\clearpage
\appendix
\renewcommand\thefigure{A\arabic{figure}}
\setcounter{figure}{0}
\renewcommand\thetable{A\arabic{table}}
\setcounter{table}{0}
\renewcommand\theequation{A\arabic{equation}}
\setcounter{equation}{0}
\pagenumbering{arabic}
\renewcommand*{\thepage}{A\arabic{page}}
\setcounter{footnote}{0}

\section{Additional qualitative results}

\begin{figure*}[t!]
    \centering
    \includegraphics[width=\linewidth]{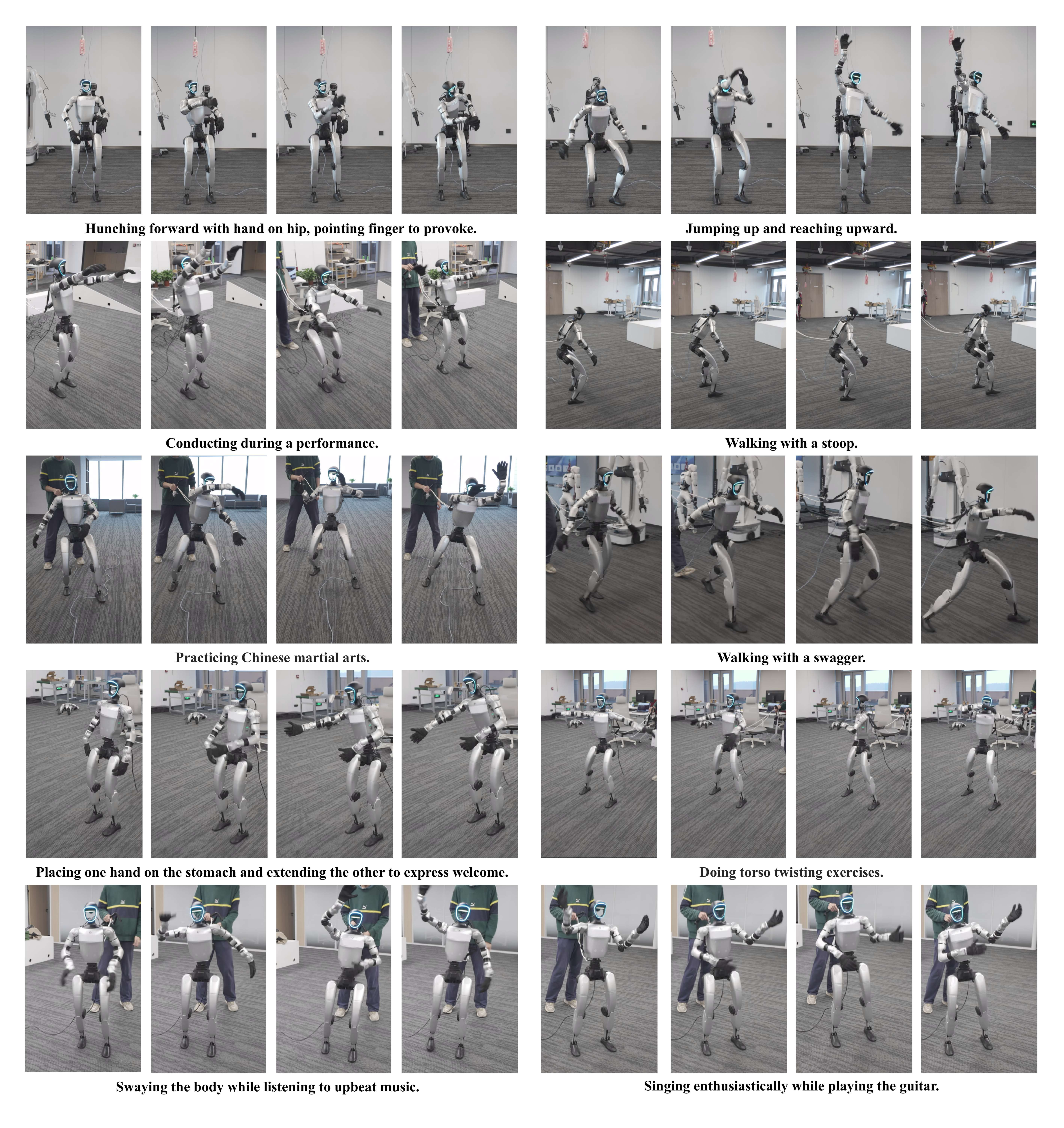}
    \caption{\textbf{Qualitative results of language-conditioned humanoid control}.}
    \label{fig:supp:text2motion_qual}
\end{figure*}

\begin{figure*}[t!]
    \centering
    \includegraphics[width=\linewidth]{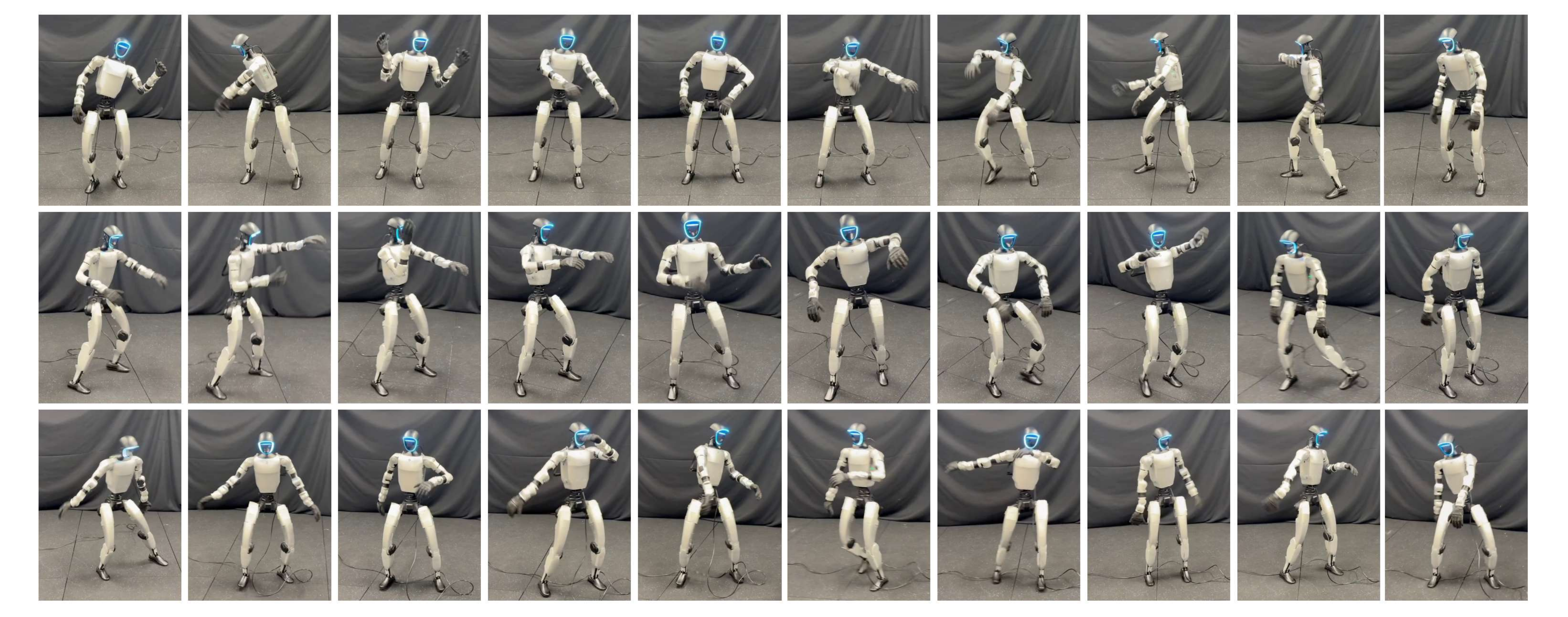}
    \caption{\textbf{Qualitative results of music-conditioned humanoid control}.}
    \label{fig:supp:music2motion_qual}
\end{figure*}

\begin{figure*}[t!]
    \centering
    \includegraphics[width=\linewidth]{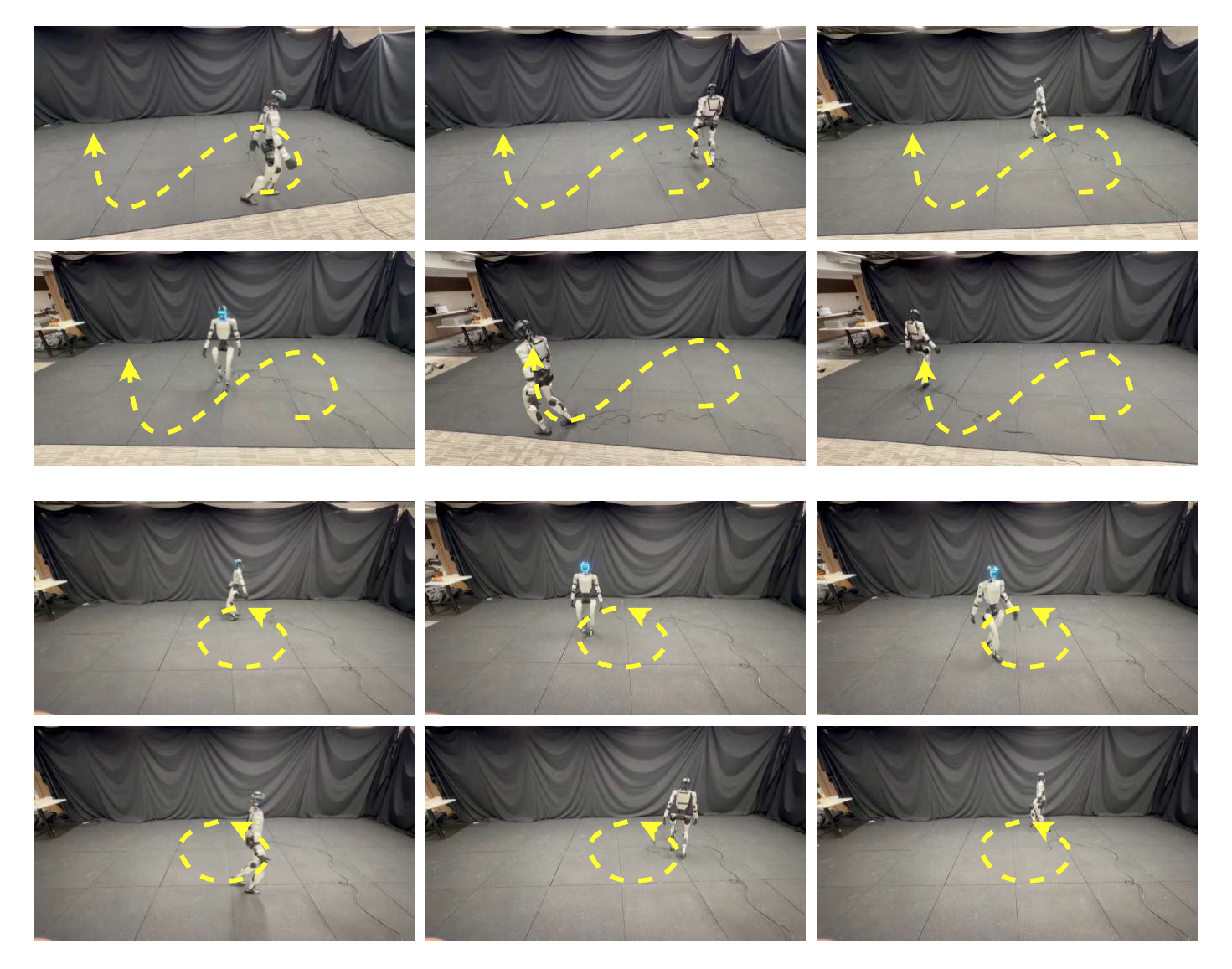}
    \caption{\textbf{Qualitative results of trajectory-conditioned humanoid control}.}
    \label{fig:supp:traj2motion_qual}
\end{figure*}

\begin{figure*}[t!]
    \centering
    \includegraphics[width=\linewidth]{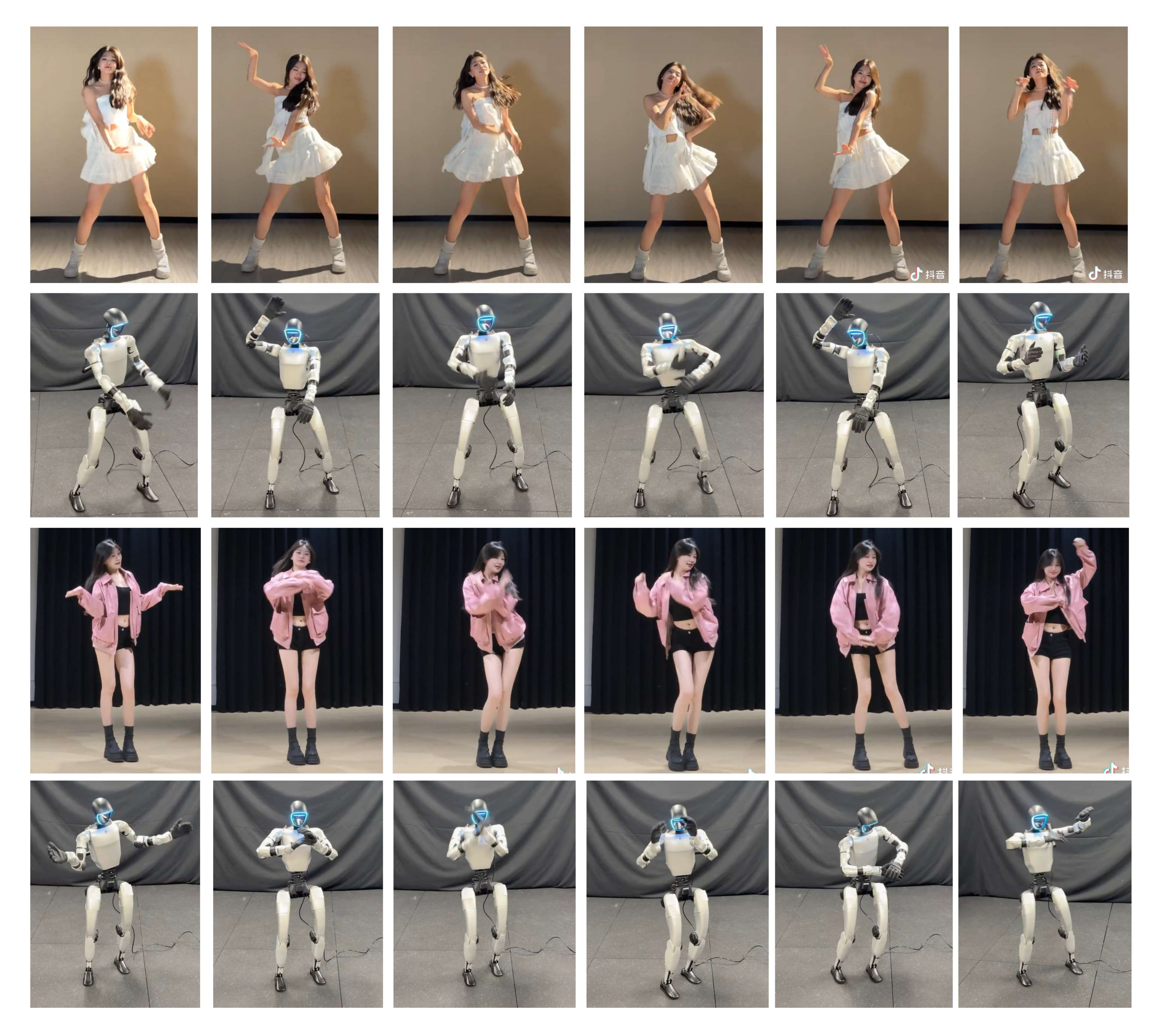}
    \caption{\textbf{Qualitative results of human-to-humanoid motion control}.}
    \label{fig:supp:gvmhr2motion_qual}
\end{figure*}

We present extensive qualitative results of real-world experiments across different modalities: language-to-motion (\cref{fig:supp:text2motion_qual}), music-to-motion (\cref{fig:supp:music2motion_qual}), trajectory-to-motion (\cref{fig:supp:traj2motion_qual}), and human-to-humanoid motion (\cref{fig:supp:gvmhr2motion_qual}). Additionally, we show examples of cross-modal control in \cref{fig:supp:cross_modal2motion_qual}. These results demonstrate that \model possesses diverse instruction following capabilities and exhibits robustness in real-world task execution. We highly recommend readers to view our \href{https://jnnan.github.io/uniact/}{\textit{project website}} for a richer visualization of \model's performance across diverse scenarios.

\section{Implementation details}

\subsection{Tokenization}\label{sec:supp:implementation_tokenization}

\paragraph{Network design} The model employs a convolutional autoencoder architecture with \acf{fsq}~\cite{mentzer2023finite} for discrete latent representation learning. The encoder-decoder framework processes sequences of shape (B, N, D) where B is batch size, N is sequence length, and D is feature dimension.

The encoder consists of an input projection layer followed by two conv blocks with increasing channel capacity. Each conv block contains a strided 1D convolution with kernel size 7 and a residual block. The residual blocks combine 1D convolutions with GroupNorm and per-position MLPs that apply LayerNorm and GELU activations. The encoder reduces the sequence length by a factor of 2 (from N to N/2) while projecting to a 5-dimensional latent space.

The decoder mirrors the encoder architecture with transposed 1D convolutions for upsampling. It takes the quantized latents and upsamples them through upsampling blocks, doubling the temporal resolution while reducing channel dimensions. The final projection layer reconstructs the original input dimension. The model uses hidden dimensions of 256 channels in intermediate layers, with expansion factors in the residual blocks for increased representational capacity. All convolution operations use appropriate padding to maintain temporal alignment throughout the network.

In ablation studies described in \cref{sec:exp}, we evaluated two architectural modifications to understand the importance of codebook capacity and temporal resolution. As shown in \cref{tab:multimodal_control,tab:motion_tracking}, reducing the codebook size to 0.25× by decreasing quantization levels resulted in performance degradation, indicating that sufficient discrete representation capacity is crucial for capturing the data distribution. Similarly, increasing the downsampling rate to 4 also decreased performance, demonstrating that maintaining adequate temporal resolution in the latent space is essential for accurate reconstruction. Excessive temporal resolution increases computational latency during motion generation. To balance motion quality with real-time performance, we adopt a downsampling rate of 2.

\paragraph{Codebook usage}
\ac{fsq} adapts its codebook size to match the complexity of different modalities. For robot motion, the \ac{fsq} quantizer discretizes the continuous degrees of freedom using predefined quantization levels [8, 8, 8, 6, 5] across 5 latent dimensions, yielding a codebook of 8×8×8×6×5=15,360 discrete codes. This larger codebook captures the nuanced variations in joint configurations and dynamic movements required for expressive robot control. In contrast, music representations utilize 6,144 codes to encode audio features, while trajectory data requires only 60 codes due to its lower-dimensional complexity and smoother spatial characteristics.

Our approach directly learns robot motion in the robot's native kinematic space, eliminating the need for post-processing retargeting steps. This end-to-end learning strategy contrasts with conventional pipelines that first generate human motion using human models (such as SMPL-X~\cite{pavlakos2019expressive}), then apply optimization-based retargeting to map the motion onto humanoid robots' workspace. Such two-stage approaches introduce cumulative errors from body proportion mismatches, joint limit violations, and the fundamental differences between human and robot morphology. Our quantitative results demonstrate that direct robot motion embedding achieves superior tracking accuracy and naturalness compared to retargeting-based methods, while also reducing computational overhead by avoiding the intermediate human representation and complex optimization procedures required for motion transfer.

The unified codebook is partitioned to accommodate multiple modalities while preserving the original text vocabulary. As listed in \cref{tab:supp:codebook_allocation}, we allocate non-overlapping index ranges for each modality: indices 0–130,076 are reserved for the original text tokens, ensuring compatibility with pretrained language models. Indices 130,077–130,078 correspond to the special tokens \texttt{<SOM>} (start of motion) and \texttt{<EOM>} (end of motion). Subsequently, indices 130,079–145,438 are designated for robot motion codes, 145,439–151,582 represent music tokens, and indices 151,583–151,642 encode trajectory representations. This design enables seamless multi-modal generation within a single autoregressive framework.

\begin{table}[ht!]
    \centering
    \small
    \setlength{\tabcolsep}{3pt}
    \caption{\textbf{Codebook allocation across modalities.}}
    \label{tab:supp:codebook_allocation}
    \begin{tabular}{lcc}
        \toprule
        Modality & Index Range & Codebook Size \\
        \midrule
        Text & 0 -- 130,076 & 130,077 \\
        \texttt{<SOM>} and \texttt{<EOM>} & 130,077 -- 130,078 & 2 \\
        Robot Motion & 130,079 -- 145,438 & 15,360 \\
        Music & 145,439 -- 151,582 & 6,144 \\
        Trajectory & 151,583 -- 151,642 & 60 \\
        \bottomrule
    \end{tabular}%
\end{table}

\subsection{Motion tracking}\label{sec:supp:motion-tracking}

We implement our motion tracking pipeline using PPO~\cite{schulman2017proximal} to train policies on the Unitree G1 humanoid robot. The framework processes hours-long reference motions from the \dataset dataset using a unified set of hyperparameters across all motions.

We adapt the motion tracking policy from BeyondMimic by removing global orientation from the reference motion observation, allowing the policy to focus on relative joint configurations:
\begin{equation}
    \mathbf{o}_t = [\mathbf{q}_t^\text{ref}, \dot{\mathbf{q}}_t^\text{ref}, \mathbf{q}_t^\text{curr}, \dot{\mathbf{q}}_t^\text{curr}, \mathbf{g}_t, \mathbf{\omega}_t, \mathbf{a}_{t-1}],
\end{equation}
where $\mathbf{q}_t^\text{ref}$ and $\dot{\mathbf{q}}_t^\text{ref}$ are the decoded reference position and velocity, $\mathbf{q}_t^\text{curr}$ and $\dot{\mathbf{q}}_t^\text{curr}$ are the current robot joint states, $\mathbf{g}_t$ represents the gravity vector in the robot frame, $\mathbf{\omega}_t$ represents the angular velocity, and $\mathbf{a}_{t-1}$ represents the robot action at previous timestep. We use asymmetric actor-critic training where the critic additionally receives per-body relative poses for direct Cartesian error estimation.

The tracker outputs joint target positions $\mathbf{p}_t$ that drive the robot to follow the generated motion while maintaining balance and physical feasibility. These are converted to torques through PD control:
\begin{equation}
    \boldsymbol{\tau}_j = k_{p,j}(\mathbf{p}_{t,j} - \mathbf{q}_{t,j}^\text{curr}) - k_{d,j}\dot{\mathbf{q}}_{t,j}^\text{curr},
\end{equation}
where $\boldsymbol{\tau}_j$ is the derived torque applied to each robot joint.

Joint stiffness and damping follow:
\begin{equation}
    k_{p,j} = I_j \omega_n^2, \quad k_{d,j} = 2 I_j \zeta \omega_n,
\end{equation}
with natural frequency $\omega_n = 10$~Hz, overdamped ratio $\zeta = 2$, and reflected inertia $I_j = k_{g,j}^2 I_{\text{motor},j}$.

\paragraph{Reward design}
The reward function primarily consists of tracking rewards that measure motion accuracy and penalty terms for physical constraints. The tracking rewards include matching the root orientation, body link positions and orientations relative to the root frame, and linear/angular velocities of all tracked links to ensure accurate motion reproduction. Key penalties include action rate regularization to reduce jitter, joint limit violations to prevent hardware damage, and undesired contacts to avoid collisions between non-end-effector body parts and the environment. For more details, please refer to \cref{tab:supp:beyondmimic}.

\paragraph{Domain randomization}
We randomize ground friction $\mu \sim \mathcal{U}(0.5, 1.5)$, joint offsets $\Delta q_j \sim \mathcal{U}(-0.05, 0.05)$ rad, and center of mass position $\Delta \mathbf{p}_{\text{CoM}} \sim \mathcal{U}(-0.02, 0.02)$ m.

\paragraph{Training configuration}
Networks use MLPs with [2048, 1024, 512, 256, 128] hidden units and ELU activations. PPO training uses clip range 0.2, learning rate $1 \times 10^{-3}$, discount $\gamma = 0.99$, GAE $\lambda = 0.95$, with 16,384 environments and control frequency 200~Hz (decimation=4).

\paragraph{Deployment}
Policies are exported as PyTorch (.pt) files and deployed at 50~Hz. Each inference step takes about 7ms on an RTX 4090 GPU.

\begin{table}[t!]
    \centering
    \small
    \setlength{\tabcolsep}{3pt}
    \caption{\textbf{Reward components and weights for OmniH2O~\cite{he2024omnih2o}.}}
    \label{tab:supp:omnih2o}
    \begin{tabular}{ccc}
        \toprule
        \textbf{Term} & \textbf{Expression} & \textbf{Weight} \\
        \midrule
        \multicolumn{3}{c}{\textbf{Penalty}} \\
        \midrule
        DoF position limits & $\mathds{1}(d_t \notin [q_{\min}, q_{\max}])$ & -1500 \\
        Termination & $\mathds{1}_{\text{termination}}$ & -250 \\
        \midrule
        \multicolumn{3}{c}{\textbf{Regularization}} \\
        \midrule
        Lower-body action rate & $\|a_t^{\text{lower}} - a_{t-1}^{\text{lower}}\|_2^2$ & -2 \\
        Upper-body action rate & $\|a_t^{\text{upper}} - a_{t-1}^{\text{upper}}\|_2^2$ & -4 \\
        Torque & $\|\tau_t\|$ & -0.0005 \\
        Slippage & $\|v_t^{\text{feet}}\|_2^2 \times \mathds{1}(F_{\text{feet}} \geq 1)$ & -5 \\
        \midrule
        \multicolumn{3}{c}{\textbf{Task Reward}} \\
        \midrule
        DoF position & $\exp(-0.25\|d_t - \tilde{d}_t\|_2)$ & 100 \\
        DoF velocity & $\exp(-0.25\|\dot{d}_t - \dot{\tilde{d}}_t\|_2^2)$ & 20 \\
        Body position & $\exp(-0.5\|p_t - \tilde{p}_t\|_2^2)$ & 50 \\
        Body rotation & $\exp(-0.1\|\theta_t \ominus \tilde{\theta}_t\|)$ & 50 \\
        Body velocity & $\exp(-10.0\|v_t - \tilde{v}_t\|_2)$ & 20 \\
        Body angular velocity & $\exp(-0.01\|\omega_t - \tilde{\omega}_t\|_2)$ & 20 \\
        \bottomrule
    \end{tabular}
\end{table}

\begin{table}[t!]
    \centering
    \small
    \setlength{\tabcolsep}{3pt}
    \caption{\textbf{Reward components and weights for BeyondMimic~\cite{liao2025beyondmimic}.}}
    \label{tab:supp:beyondmimic}
    \resizebox{\linewidth}{!}{%
        \begin{tabular}{ccc}
            \toprule
            \textbf{Term} & \textbf{Expression} & \textbf{Weight} \\
            \midrule
            \multicolumn{3}{c}{\textbf{Penalty}} \\
            \midrule
            Action rate & $\|{a}_t-{a}_{t-1}\|_2^2$ & -0.1 \\
            Joint limit & $\sum_{j} \mathds{1}[{q}_{t,j} \notin [{q}_{t,j}^{\text{min}}, {q}_{t,j}^{\text{max}}]]$ & -10.0 \\
            Undesired contacts & $\sum_{c \notin \{\text{ankles, wrists}\}} \mathds{1}[\|{F}_c\|>1.0\text{N}]$ & -0.1 \\
            \midrule
            \multicolumn{3}{c}{\textbf{Task Reward}} \\
            \midrule
            Root orientation & $\exp\!\big(- \|{o}^p_{t,r}-{o}^g_{t,r}\|_2^2 / 0.4^2\big)$ & 0.5 \\
            Body link pos (rel.) & $\exp\!\Big(-\tfrac{1}{|\mathcal{B}|}\!\sum_{b\in\mathcal{B}}\|{p}^{p,\text{rel}}_{t,b}-{p}^{g,\text{rel}}_{t,b}\|_2^2 / 0.3^2\Big)$ & 1.0 \\
            Body link ori (rel.) & $\exp\!\Big(-\tfrac{1}{|\mathcal{B}|}\!\sum_{b\in\mathcal{B}} \|{o}^{p,\text{rel}}_{t,b}-{o}^{g,\text{rel}}_{t,b}\|_2^2 / 0.4^2\Big)$ & 1.0 \\
            Body link lin. vel & $\exp\!\Big(-\tfrac{1}{|\mathcal{B}|}\!\sum_{b\in\mathcal{B}}\|{v}^p_{t,b}-{v}^g_{t,b}\|_2^2 / 1.0^2\Big)$ & 1.0 \\
            Body link ang. vel & $\exp\!\Big(-\tfrac{1}{|\mathcal{B}|}\!\sum_{b\in\mathcal{B}}\|{\omega}^p_{t,b}-{\omega}^g_{t,b}\|_2^2 / 3.14^2\Big)$ & 1.0 \\
            \bottomrule
        \end{tabular}
    }%
\end{table}

\section{Experimental details}

\subsection{Baseline methods}\label{sec:supp:baseline_adaptation}

We apply OmniH2O~\cite{he2024omnih2o}, UH-1~\cite{mao2024learning}, and LangWBC~\cite{shao2025langwbc} as our baselines. For OmniH2O, we adapt the original settings from the H1 robot to G1, with the adapted parameters shown in \cref{tab:supp:omnih2o}. We leverage Qwen2.5-3B~\cite{bai2025qwen25vltechnicalreport} throughout these comparisons.

For UH-1, we adopt the end-to-end configuration where the method directly generates robotic actions $A_\text{robot}$ from text input for open-loop control (text-to-action). To ensure fair comparison with our approach, we train the control policy using the same architecture and training procedure as the motion tracker in \model. Specifically, the policy network learns to map text embeddings directly to joint-level control commands without intermediate motion representation. The parameters are adapted from the original Unitree H1-2 configuration to accommodate the G1 robot.

For LangWBC, we reimplement the method using the BeyondMimic tracker as the teacher policy with reward formulation in \cref{tab:supp:beyondmimic}. We maintain the original CVAE architecture and DAgger training but scale to 4,096 parallel environments. Unlike the original implementation that tracks only root displacement, we extend the loss computation to whole-body tracking across all keypoints, enabling more accurate motion reproduction on the G1 robot.

We employ the motion diffusion model (MDM)~\cite{mdm} as the motion generator for two-stage baseline methods. The MDM model was trained using motions from the \dataset training set, which were converted to SMPL-X format~\cite{pavlakos2019expressive} for compatibility.

\subsection{Evaluation metrics}\label{sec:supp:metric}

We leverage multiple metrics to measure \model's performance systematically. Following~\citet{humanml3d} and~\citet{mdm}, we employ \ac{fid}, Diversity, MM-Dist, and R-precision. \ac{fid} assesses motion quality by measuring the distance between the distribution of generated motions and training motion data. Diversity measures the generation diversity by calculating the distance between different motion features. MM-Dist and R-precision evaluate the alignment between generated motions and their corresponding multimodal instructions. For text-to-motion, we compute MM-Dist and R-precision using text features corresponding to each motion. For trajectory-to-motion and music-to-motion, we use ground truth motion features to represent instruction features. Following TMR~\cite{tmr}, we train motion and text encoders using contrastive learning on the \dataset dataset to extract motion and text features.

\Ac{rmse} measures the distance between the robot motion's root and the trajectory in the trajectory-to-motion task. We propose the genre metric to measure the diversity of dance motions generated from music of the same style in the music-to-motion task. It calculates the distance between motion features generated from music of the same genre. Success rate measures the percentage of trials where the robot completes the task without falling or significant deviation from instructions. For text-to-motion, a trial is considered successful if the robot does not fall and the MPJPE between the robot motion and ground truth motion is below 0.8m. For trajectory-to-motion, a trial is considered successful if the robot does not fall and the \ac{rmse} is below 1.0m. For music-to-motion, a trial is considered successful if the robot does not fall.

\begin{figure*}[t!]
    \centering
    \small
    \setlength{\tabcolsep}{3pt}
    \begin{subfigure}[t]{.5\linewidth}
        \includegraphics[width=\linewidth]{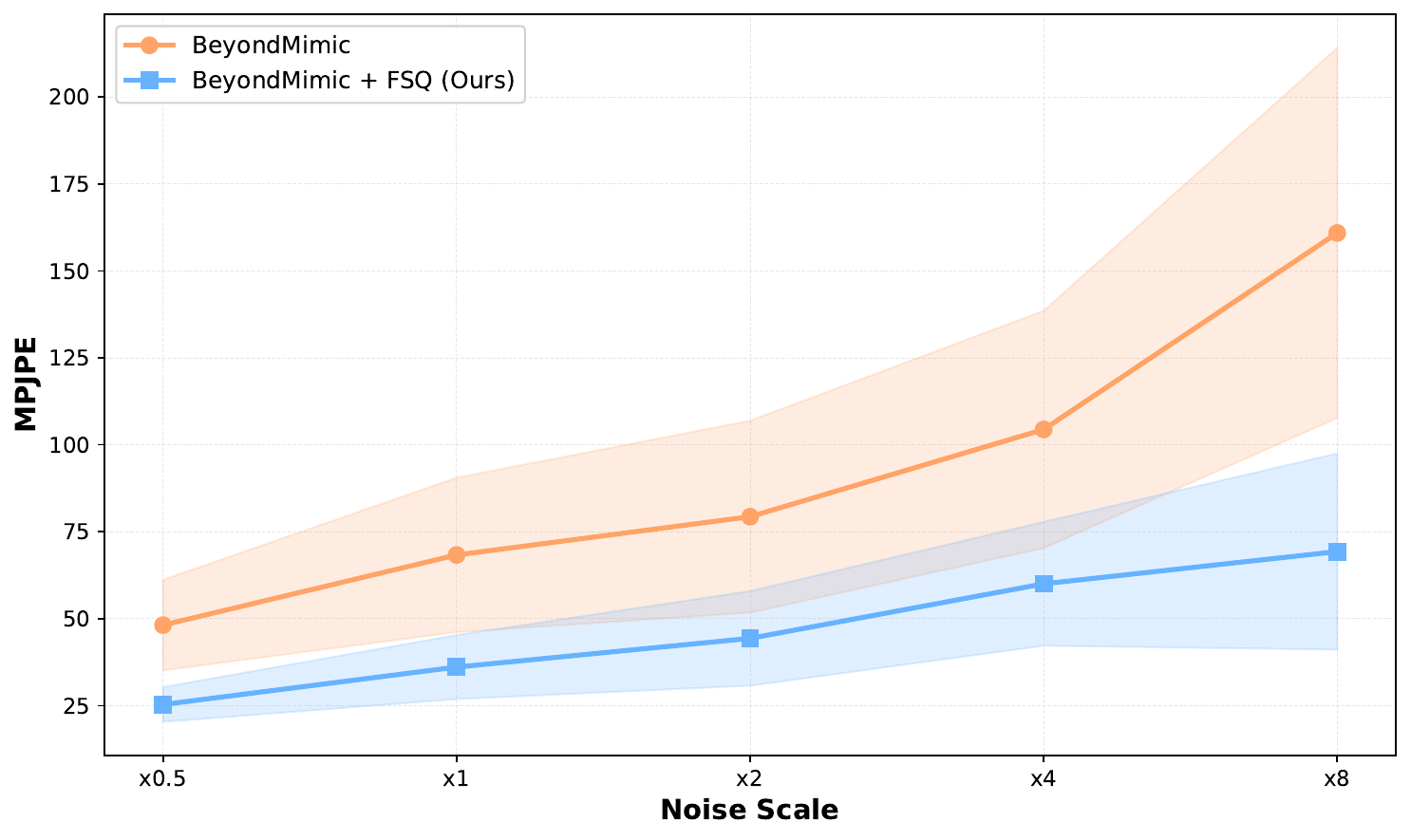}
    \end{subfigure}%
    \begin{subfigure}[t]{.5\linewidth}
        \includegraphics[width=\linewidth]{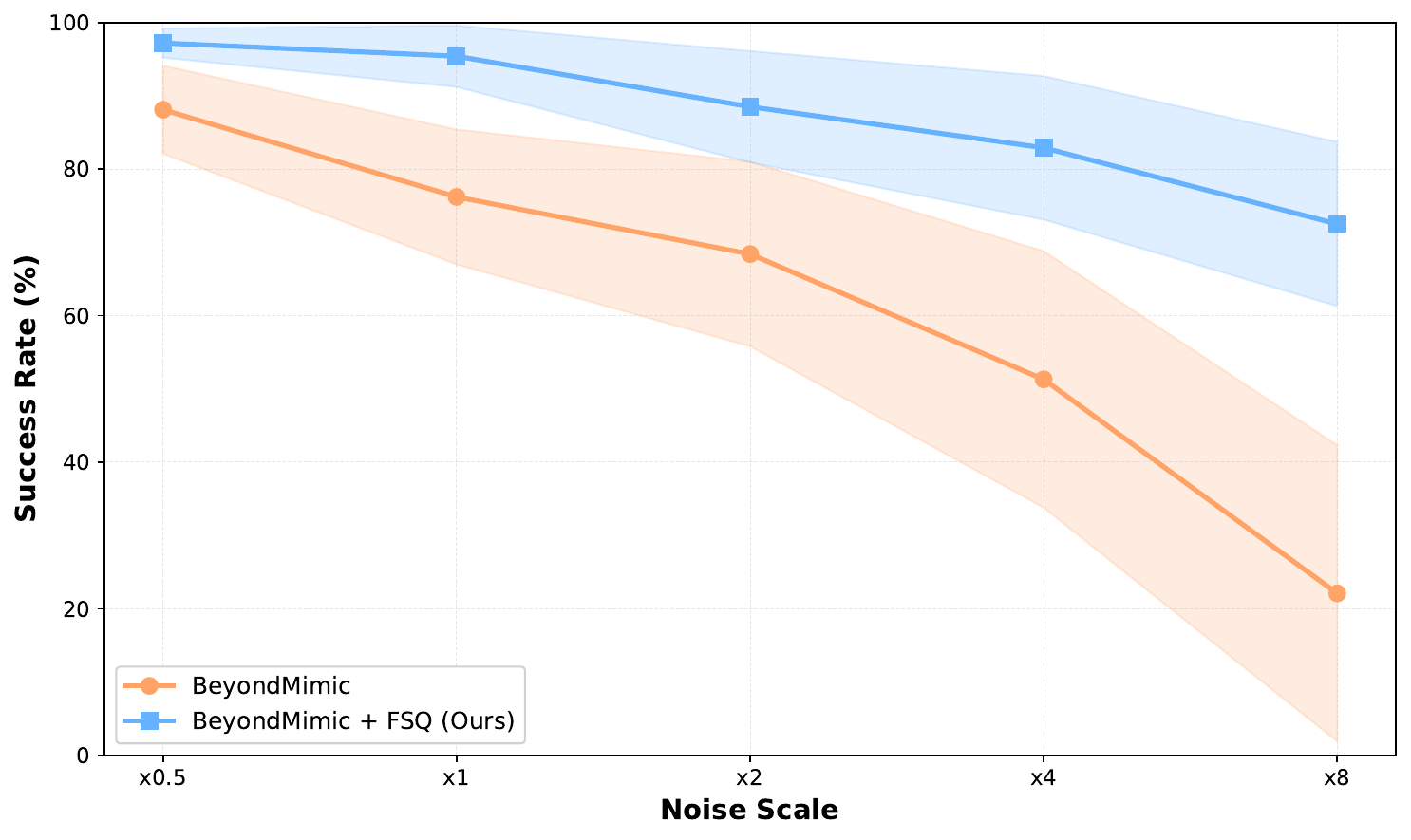}
    \end{subfigure}
    \caption{\textbf{Performance comparison between BeyondMimic and BeyondMimic + \acs{fsq} (ours) under different noise scales.} (left) MPJPE comparison. (right) Success rate comparison. As the noise scale progressively increases, BeyondMimic exhibits significant performance degradation, whereas BeyondMimic + \acs{fsq} (ours) maintains relatively low MPJPE and high success rates even at the maximum noise scale (×8), which demonstrates that our quantization approach effectively constrains motions to feasible spaces, enhancing robustness.}
    \label{fig:supp:fsq_tracking_performance_comparison}
\end{figure*}

\section{The \dataset dataset}\label{sec:supp:dataset}

\subsection{Trajectory expansion with motion-matching}
\begin{figure}[t!]
    \centering
    \includegraphics[width=\linewidth]{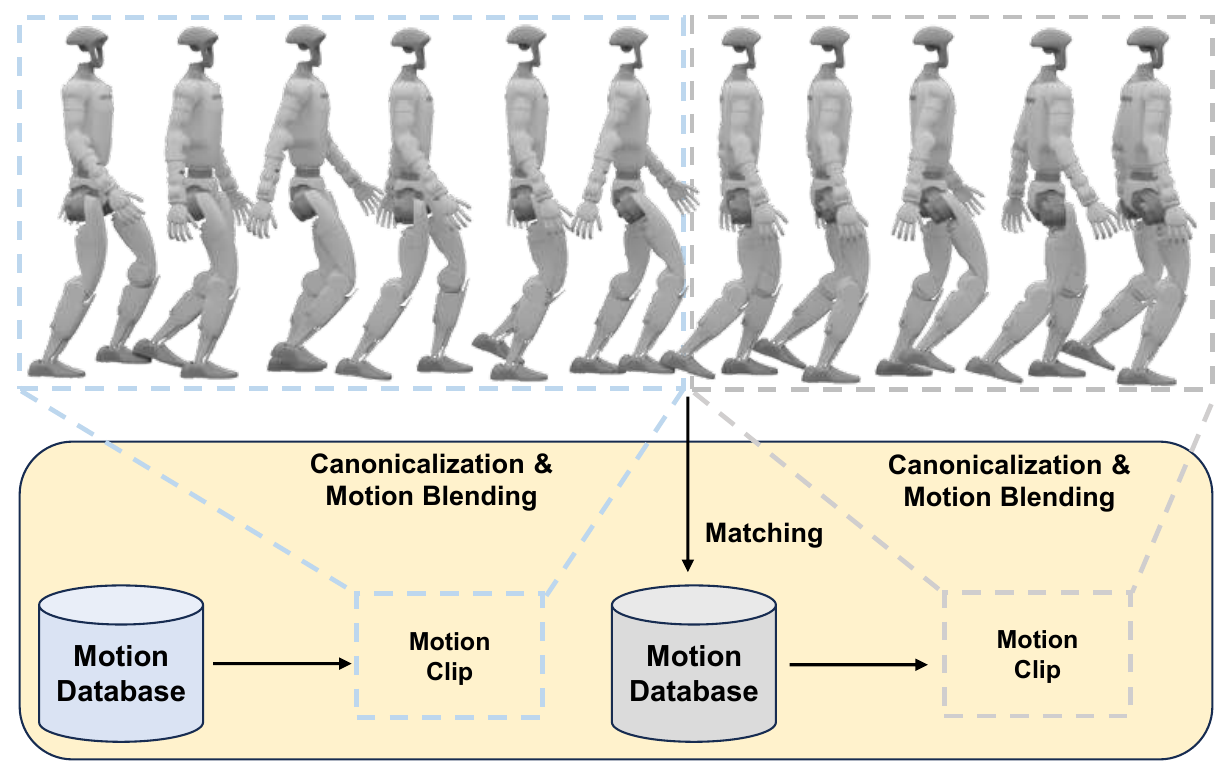}
    \caption{\textbf{Overview of motion-matching-based walking motion expansion}. To expand the human walking trajectories in the \dataset dataset from 20 minutes to over 10 hours, we employ a motion-matching-based augmentation approach. The original motion sequences are segmented into short clips, with each frame represented by features capturing pose, velocity, and future trajectory information. During synthesis, the system continuously searches for optimal clip transitions based on feature similarity and blends them smoothly.}
    \label{fig:supp:motionmatching}
\end{figure}

We collected approximately 20 minutes of human walking motion data on flat terrain using the OptiTrack MoCap system. To obtain paired data of arbitrary trajectories and walking motions, we augmented this dataset using motion matching~\cite{clavet2016motionmatching}, expanding the walking motion data from 20 minutes in the \dataset dataset to over 10 hours. This motion-matching-based augmentation pipeline preserves natural locomotion characteristics while following diverse, non-repetitive trajectories. The motion matching pipeline is present in \cref{fig:supp:motionmatching}.

\paragraph{Feature extraction}
In motion matching, each frame is represented by a comprehensive feature vector containing root position and orientation, relative joint positions for lower-body joints (foot and ankles), root and foot velocities, future trajectory points sampled at 20, 40, and 60 frames ahead, and gait phase information derived from foot contact patterns. These features capture both the instantaneous pose and the movement intention, enabling accurate matching between motion clips.

\paragraph{Segmentation process}
The original 20-minute sequences are segmented into overlapping clips of 2-4 seconds, with cut points determined by gait cycle boundaries at heel-strike events and steady-state walking phases where acceleration is minimal. Each segment maintains at least two complete gait cycles and is annotated with metadata, including walking speed and turning angle. The overlapping strategy ensures multiple transition possibilities from each clip.

\paragraph{Matching and synthesis}
During synthesis, the system searches for optimal transitions when approaching the last 10 frames of the current clip. The matching cost function evaluates feature similarity through weighted Euclidean distance, velocity consistency, trajectory alignment, and foot contact states. To ensure diversity, we implement a history buffer that penalizes recently used clips and probabilistically select from the top-5 candidates rather than always choosing the best match.

Transitions are blended over 10 frames using spherical linear interpolation for rotations and cubic interpolation for positions. Post-processing applies inverse kinematics to maintain foot contacts during stance phases and smooths trajectories to eliminate artifacts. The pipeline continuously monitors quality metrics, including foot sliding and acceleration discontinuities, automatically filtering problematic sequences.

This approach successfully synthesizes over 10 hours of varied, natural walking motions from the original 20-minute MoCap data, providing sufficient diversity for training trajectory following.

\subsection{Text-to-motion from MoCap}

The \dataset dataset comprises over 10,000 expressive motion sequences with meticulously curated motion descriptions. We systematically cover 1,400 common action verbs, representing 1.8× the vocabulary coverage of HumanML3D~\cite{humanml3d} dataset. For each verb, we compose contextually appropriate sentences that naturally incorporate the verb as the primary action descriptor. High-frequency verbs receive additional instances to reflect their prevalence in real-world applications, with frequently used actions like `walk,' `run,' and `turn' appearing in multiple distinct motion descriptions to ensure comprehensive coverage of common movements.

We employed 10 professional MoCap actors (4 male, 6 female), carefully selected for similar body proportions to minimize morphological variation in the dataset. All actors possess foundational dance training, ensuring they can execute movements with proper body control and spatial awareness. During capture sessions, actors were instructed to perform motions with slight expressiveness beyond everyday movements, adding subtle stylistic flourishes and dynamic variations that enhance the dataset's richness while maintaining naturalness. This directive encouraged actors to interpret each motion with personality rather than mechanical execution, resulting in more engaging and diverse performances that better reflect how humans naturally move when given verbal instructions.

\section{Robustness analysis under input corruption}\label{sec:supp:robustness_analysis}

To evaluate whether our discrete representation can effectively refine noisy inputs, we introduce Gaussian noise and temporal jitter to \dataset motions to simulate low-quality input data. The noise model mimics low-quality MoCap data by incorporating three primary components that represent common tracking artifacts. Given an input motion sequence $\mathbf{X} \in \mathbb{R}^{N \times D}$ with $N$ frames and $D = 29$ degrees of freedom, we generate the noisy output as:
\begin{equation}
    \tilde{\mathbf{X}} = \mathbf{X} + \mathbf{n}_{\text{base}} + \mathbf{n}_{\text{burst}} + \mathbf{n}_{\text{jitter}},
\end{equation}
where $\mathbf{n}_{\text{base}} \sim \mathcal{N}(0, \sigma_{\text{base}}^2\mathbf{I})$ with $\sigma_{\text{base}} = 0.01$ represents baseline sensor noise inherent in all MoCap systems. The burst noise $\mathbf{n}_{\text{burst}}$ simulates temporary tracking failures that occur with probability $p_{\text{burst}} = 0.05$ per frame, affecting random subsets of dimensions with temporally correlated Gaussian perturbations $\mathcal{N}(0, \sigma_{\text{burst}}^2\mathbf{I})$ where $\sigma_{\text{burst}} = 0.1$ over durations of 8-20 frames. The jitter noise $\mathbf{n}_{\text{jitter}}$ models sudden tracking spikes that occur with probability $p_{\text{jitter}} = 0.001$ per dimension per frame, introducing large-magnitude disturbances drawn from $\mathcal{N}(0, \sigma_{\text{jitter}}^2\mathbf{I})$ where $\sigma_{\text{jitter}} = 0.3$. After processing these corrupted motions through our discrete representation, we compare the tracking performance against the original unperturbed references to assess the quantization's denoising capabilities. 

As shown in \cref{tab:motion_tracking}, all quantization settings achieve significantly better tracking performance compared to the baseline tracking method, where corrupted motions are directly fed into the BeyondMimic motion tracker. This demonstrates that the discrete token representation effectively filters out noise by constraining motions to a finite, well-explored space of learned motion patterns. The quantization process inherently rejects out-of-distribution perturbations and projects noisy inputs onto the manifold of plausible human motions, thereby significantly enhancing tracking robustness even under substantial input corruption.

To further validate the advantages of token-based motion representation, we evaluated the performance of both BeyondMimic~\cite{liao2025beyondmimic} and BeyondMimic + \ac{fsq}~\cite{mentzer2023finite} (ours) under varying noise scales. As shown in \cref{fig:supp:fsq_tracking_performance_comparison}, as the noise scale progressively increases, BeyondMimic exhibits significant performance degradation, whereas BeyondMimic + \ac{fsq} (ours) maintains relatively low MPJPE and high success rates even at the maximum noise scale (×8). This demonstrates that our method can effectively restore the quality of out-of-distribution low-quality data to a certain extent, thereby providing more feasible reference motions for the motion tracking policy.

\section{Compositional cross-modal control}\label{sec:supp:cross-modal-control}

\model extends naturally to cross-modal humanoid control tasks, where robots must simultaneously process and execute instructions from multiple input modalities. This capability is exemplified by scenarios requiring the robot to perform expressive gestures, such as waving, while simultaneously following a specified trajectory. Such multi-modal control is crucial for practical deployment scenarios where robots must navigate through environments while maintaining meaningful interactions with humans.

Our approach leverages a key observation: expressive semantic actions predominantly manifest in the upper body of humanoid robots, while trajectory following primarily engages lower-body locomotion. Given this observation, we independently generate motions from language and trajectory instructions, then synthesize composed motion by combining the upper-body motion from text-conditioned generation with the lower-body motion from trajectory-conditioned generation. Each robot joint belongs to either the upper body or the lower body, with the pelvis joint serving as the boundary. The pelvis joint itself belongs to the lower body to maintain stable locomotion dynamics. The composed motion is subsequently forwarded to the motion tracking policy for real-world execution.

\begin{figure*}[t!]
    \centering
    \includegraphics[width=\linewidth]{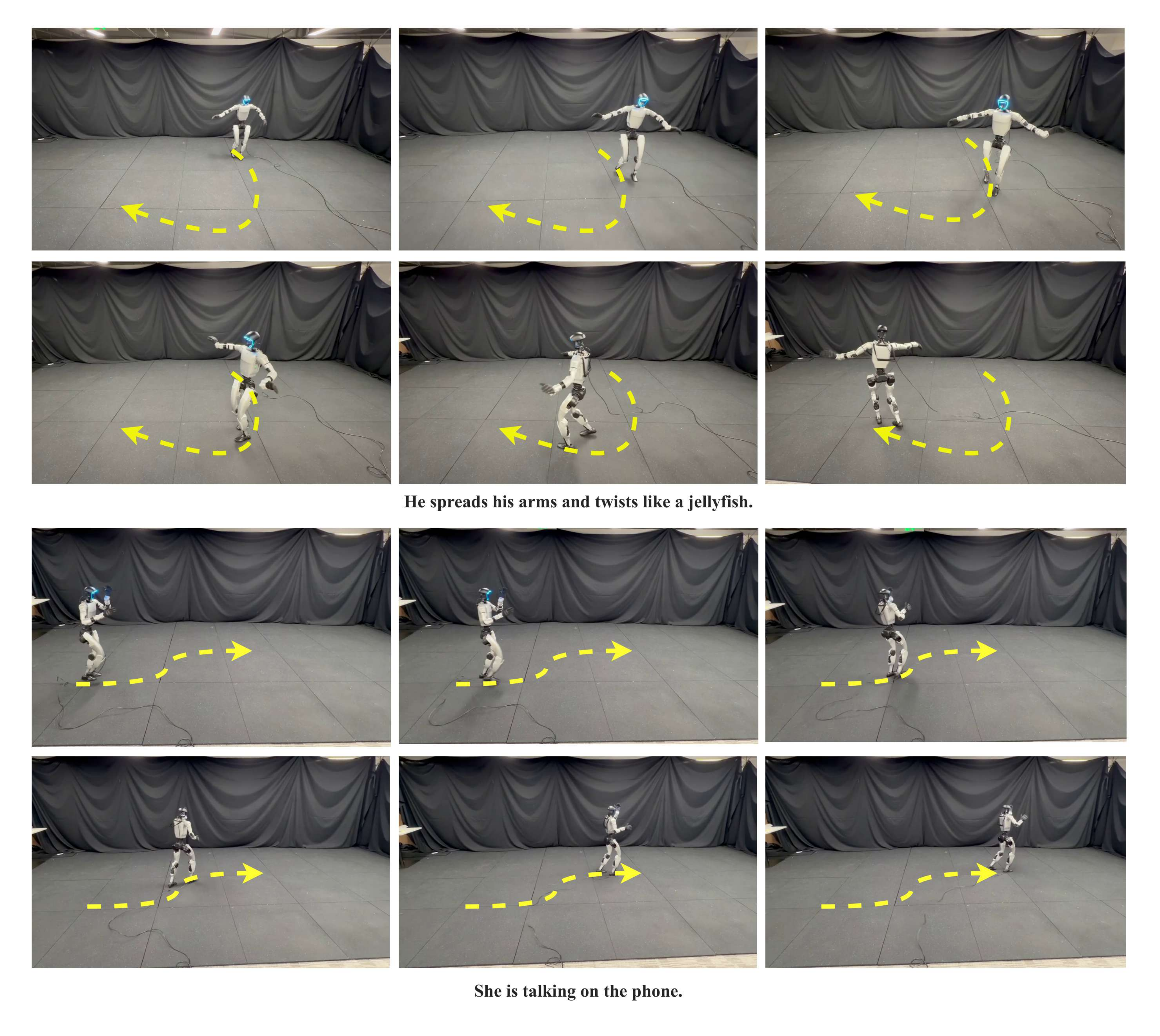}
    \caption{\textbf{Qualitative results of cross-modal humanoid control}.}
    \label{fig:supp:cross_modal2motion_qual}
\end{figure*}

To ensure robust tracking of these compositionally generated motions, we implement a fine-tuning strategy for the tracking policy. During fine-tuning, we systematically compose motions by randomly pairing lower-body segments from trajectory data with upper-body segments from other motion categories within the \dataset dataset. This augmentation strategy exposes the tracking policy to motion patterns structurally similar to those encountered during cross-modal inference, enabling the successful execution of previously unseen cross-modal compositions. \cref{fig:supp:cross_modal2motion_qual} illustrates the results from our cross-modal experiments, demonstrating coherent execution of combined instructions.

This compositional approach enables zero-shot generalization to novel motion-trajectory combinations. For instance, despite the training dataset containing only stationary versions of actions such as `making a phone call' or `playing drums', our method successfully controls the robot to perform these actions while following arbitrary trajectories. This capability significantly expands the robot's behavioral diversity without requiring exhaustive data collection for every possible combination of actions and trajectories, thereby demonstrating the scalability and practical utility of our framework for real-world humanoid deployment.
\clearpage

\clearpage

\clearpage

\clearpage

\end{document}